\documentclass[10pt,twocolumn,letterpaper]{article}

\usepackage{cvpr}              

\newcommand{\virstfull}{\textbf{VIRST} (\textbf{V}ideo-\textbf{I}nstructed \textbf{R}easoning Assistant for \textbf{S}patio-\textbf{T}emporal Segmentation)}
\newcommand{\virstabbr}{VIRST}
\newcommand{\stmodulefull}{{Spatio-Temporal Fusion (STF)}}
\newcommand{\stmoduleabbr}{STF}
\newcommand{\memorymodulefull}{{Temporal Dynamic Anchor Updater (TDAU)}}
\newcommand{\memorymoduleabbr}{{TDAU}}
\newcommand{\sttoken}{{\texttt{[ST]}}}
\newcommand{\jnf}{{$\mathcal{J}\&\mathcal{F}$}}
\newcommand{\stageone}{Alignment Stage}
\newcommand{\stagetwo}{Few-Image Prediction Stage}
\newcommand{\stagethree}{Propagation Stage}

\usepackage{url}
\usepackage{graphicx}
\usepackage{booktabs}
\usepackage{multirow}
\usepackage{makecell}
\usepackage{subcaption}
\usepackage{caption}
\usepackage{wrapfig}
\usepackage{placeins}
\usepackage{tabularx}
\usepackage[ruled,vlined]{algorithm2e}
\usepackage[accsupp]{axessibility}

\definecolor{cvprblue}{rgb}{0.21,0.49,0.74}
\usepackage[pagebackref,breaklinks,colorlinks,allcolors=cvprblue]{hyperref}

\def\paperID{} 
\def\confName{CVPR}
\def\confYear{2026}

\title{VIRST: Video-Instructed Reasoning Assistant for SpatioTemporal Segmentation}

\author{
Jihwan Hong$^{1}$ \quad Jaeyoung Do$^{1,2,}$\thanks{Corresponding author}\\
AIDAS Laboratory, $^1$IPAI \& $^2$ECE, Seoul National University \\
{\tt\small \{csjihwanh, jaeyoung.do\}@snu.ac.kr}
}

\begin{document}
\maketitle
\begin{abstract}

Referring Video Object Segmentation (RVOS) aims to segment target objects in videos based on natural language descriptions. However, fixed keyframe-based approaches that couple a vision language model with a separate propagation module often fail to capture rapidly changing spatiotemporal dynamics and to handle queries requiring multi-step reasoning, leading to sharp performance drops on motion-intensive and reasoning-oriented videos beyond static RVOS benchmarks. To address these limitations, we propose \virstfull, an end-to-end framework that unifies global video reasoning and pixel-level mask prediction within a single model. \virstabbr\ bridges semantic and segmentation representations through the \stmodulefull, which fuses segmentation-aware video features into the vision-language backbone, and employs the \memorymodulefull\ to maintain temporally adjacent anchor frames that provide stable temporal cues under large motion, occlusion, and reappearance. This unified design achieves state-of-the-art results across diverse RVOS benchmarks under realistic and challenging conditions, demonstrating strong generalization to both referring and reasoning oriented settings. The code and checkpoints are available at \url{https://github.com/AIDASLab/VIRST}.

\end{abstract}    
\section{Introduction}
\label{sec:intro}

\begin{figure}[t]
  \centering
  \setlength{\abovecaptionskip}{6pt}
  \setlength{\belowcaptionskip}{-6pt}
  \includegraphics[width=\linewidth, trim=0 20 0 5, clip]{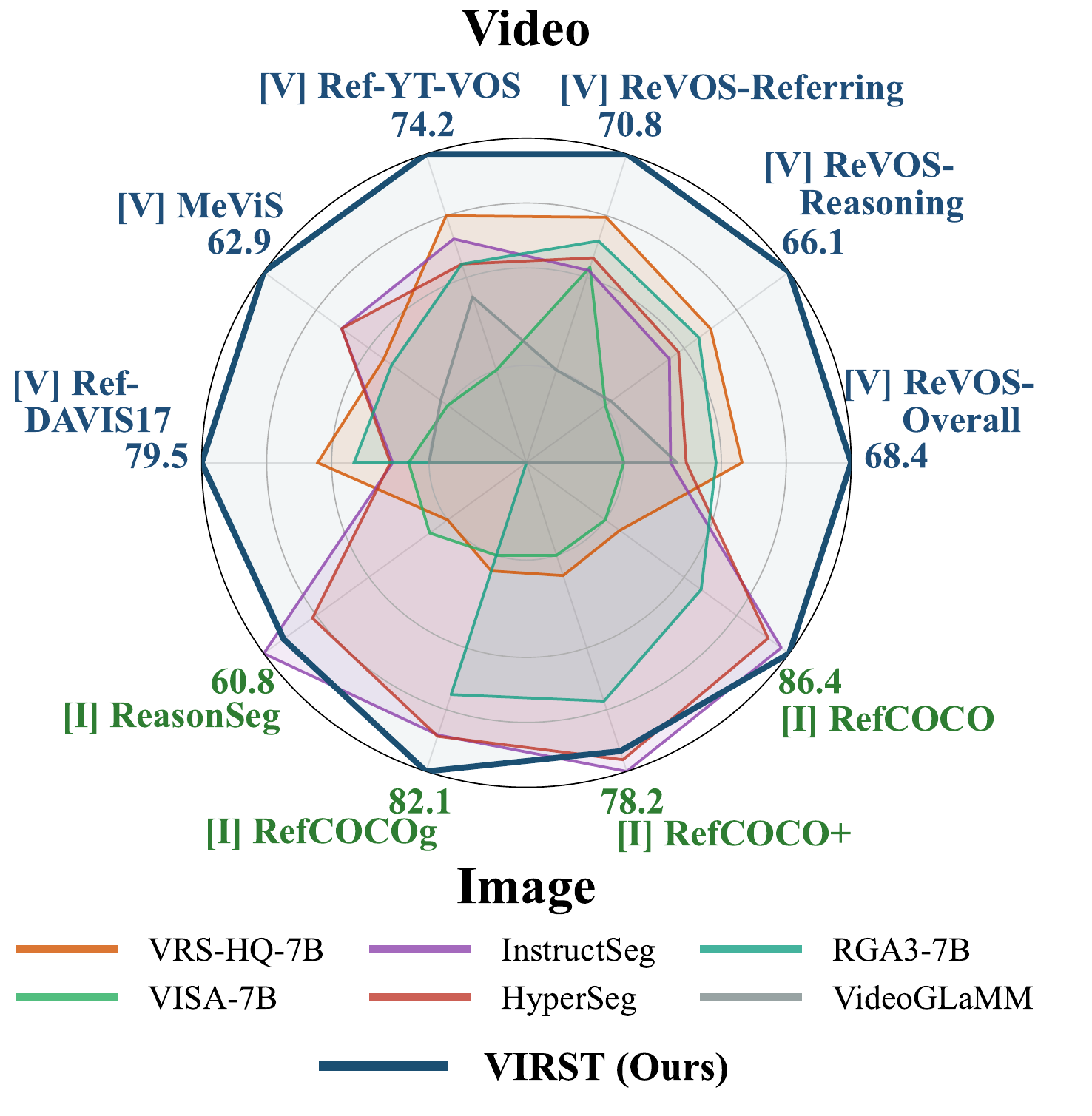}
  \caption{
    \textbf{Performance comparison with existing RVOS methods.} 
    \virstabbr{} achieves state-of-the-art results across all referring video object segmentation benchmarks, 
    while maintaining competitive performance on referring and reasoning-based image segmentation tasks.
  }
  \label{fig:main_spider}
\end{figure}

\begin{figure*}[t]
  \centering
    \setlength{\abovecaptionskip}{6pt}
    \setlength{\belowcaptionskip}{-6pt}

    \includegraphics[width=1\linewidth]{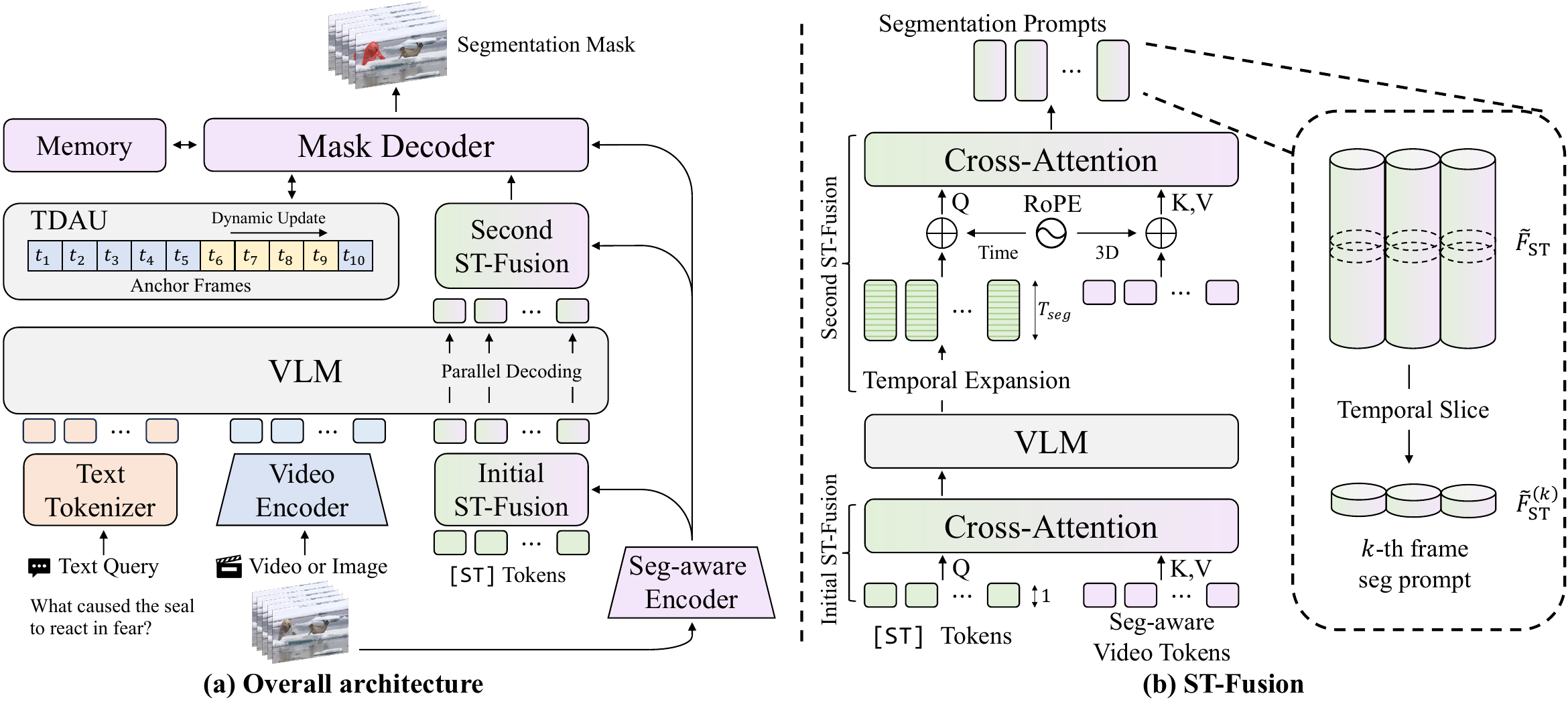} 
    \caption{
    \textbf{Overall architecture of \virstabbr{}.}
    (a) \virstabbr{} utilizes a VLM to capture global video context and identify query-aligned targets. The \stmodulefull{} fuses features from the segmentation-aware vision encoder, while the \memorymodulefull{} provides local and long-range temporal cues through a dual-track memory design.
    (b) The ST-Fusion module includes an Initial ST-Fusion stage, where the \sttoken{} tokens are fused with segmentation-aware video tokens prior to VLM processing, followed by the Second ST-Fusion stage that applies cross-attention between the temporally expanded \sttoken{} tokens and the segmentation-aware video tokens. The resulting spatiotemporal prompts are sliced to produce frame-specific segmentation prompts.
    }

    \label{fig:overall_arch}
\end{figure*}

Referring Video Object Segmentation (RVOS)~\citep{perazzi2016davis-vos,gavrilyuk2018actor,ding2023mevis,yan2024visa,seo2020urvos} seeks to segment objects in a video based on a natural-language description. This task requires far more than simply identifying the object mentioned in text: the model must perform fine-grained pixel-level understanding, precisely align language with visual concepts, and reason over complex spatial and temporal dynamics. Unlike supervised~\cite{caelles2017osvos,perazzi2016davis-vos,xu2018youtubevos,ding2023mose,ravi2024sam2}, or interactive VOS~\cite{ravi2024sam2,mei2025sami2v}, which require explicit user-provided masks or corrections, RVOS depends solely on linguistic supervision, offering a scalable and annotation-efficient pathway particularly valuable for robotics and embodied AI systems~\cite{wang2024medical-rsvis,jiang2024clipunetr,wang2022audiovisualgrounding}.

Recent progress has been driven by the integration of vision-language models (VLMs) into RVOS pipelines, where a VLM interprets the text-video input to generate segmentation masks on a given keyframe, and then a video segmentation module decodes pixel-level masks. Methods following this paradigm, such as VISA~\citep{yan2024visa} and other successors~\citep{gong2025vrshq,bai2024videolisa_reasonvos,zheng2025villa,wang2025rga3} have shown that VLMs can strengthen global reasoning and semantic grounding.

However, despite these advances, existing approaches remain constrained by three fundamental limitations. (1) \textbf{Limited keyframe selection: } The reliance on a small, fixed set of keyframes restricts temporal robustness. Real videos contain rapid appearance changes and occlusions, and using only one or two fixed keyframes frequently causes drift and large propagation errors, resulting in inaccurate segmentation results.
(2) \textbf{Dependence on external modules:} Most pipelines depend on external modules (e.g., CLIP~\citep{cherti2023openclip} or video-LLMs~\cite{li2024llama-vid,wang2024groundedvideollm}) that are not optimized for the pixel-level segmentation module. This prevents end-to-end learning, and semantic misalignment between modules often leads to brittle or inconsistent mask predictions. 
(3) \textbf{Encoder discrepancy:} Encoding videos with CLIP-based VLM backbones that focus on semantic representations often causes misalignment with segmentation modules, which rely on hierarchical architectures designed to capture fine-grained spatial information~\cite{ryali2023hiera}; while feeding the vision features of the segmentation module into the VLM can mitigate this gap, it introduces redundancy and increases computational cost.

These limitations reveal a deeper issue: existing RVOS models treat reasoning and segmentation as loosely coupled steps. As a consequence, they achieve strong performance on static or salient-object benchmarks such as Ref-YT-VOS~\citep{seo2020urvos} and Ref-DAVIS17~\citep{khoreva2018ref-davis17} (around 70~\jnf), yet drop sharply by more than 10–20 points on reasoning-intensive or motion-centric datasets such as MeViS~\citep{ding2023mevis} and ReVOS~\citep{yan2024visa}. A unified architecture capable of performing language grounding, spatio-temporal reasoning, and pixel-level segmentation in a single coherent process remained elusive.

In this work, we propose \virstfull{}, which resolves these limitations through an integrated design. \virstabbr{} unifies global video reasoning and fine-grained mask prediction within a single VLM forward pass. 
At its core is \textbf{\stmodulefull{}}, which merges dense segmentation-aware video features with semantic features to encode spatiotemporal \sttoken{} tokens that serve as compact and expressive prompts for spatiotemporal reasoning. 
This resolves the encoder discrepancy by combining the spatial fidelity of segmentation-aware features with the reasoning capacity of the VLM.

To address temporal instability, we introduce \textbf{\memorymodulefull{}}, which selects and updates multiple anchor frames to provide spatiotemporally informative cues. Rather than relying on fixed keyframes, the model generates anchor-frame candidates and dynamically selects those that are temporally close and provide spatial features for segmenting the target frame.

Finally, a carefully designed progressive training strategy ensures stable optimization, aligning semantics, spatial grounding, and temporal coherence in a stepwise manner.

With this unified formulation, \virstabbr{} bridges the long-standing gap between high-level video understanding and pixel-accurate segmentation. The model achieves state-of-the-art results across four major RVOS benchmarks, outperforming previous methods by substantial margins, particularly on reasoning-oriented datasets. Beyond videos, \virstabbr{} generalizes strongly to referring and reasoning-based image segmentation tasks, demonstrating that its spatiotemporal fusion and anchor-based reasoning mechanisms act as powerful priors for multi-modal grounding.

\section{Related Works}

\paragraph{Video Language Models.}
Early vision–language models such as BLIP-2~\cite{li2023blip2} and LLaVA~\cite{liu2023llava} established strong image–text alignment and the standard visual-encoder–LLM pipeline.
VideoChat~\cite{li2023videochat} extended this paradigm to videos using a video foundation model~\cite{wang2022internvideo}, followed by Video-ChatGPT~\cite{maaz2024videochatgpt}, Video-LLaVA~\cite{lin2024videollava}, and Chat-UniVi~\cite{jin2024chatunivi}, which introduced video-specific instruction tuning with spatiotemporal visual encoding.
Long-video models such as LLaMA-VID~\cite{li2024llama-vid} and VideoChat-Flash~\cite{li2024videochatflash}, which use token compressions schemes, further improve temporal scalability for hour-long video understanding.

\paragraph{Referring Video Object Segmentation.}
Referring Video Object Segmentation (RVOS)~\cite{gavrilyuk2018actor,khoreva2018ref-davis17,seo2020urvos,ding2023mevis,yan2024visa} aims to segment a target object in a video given a natural-language description.
Early approaches~\cite{botach2022mttr,seo2020urvos,ding2023mevis,wu2022referformer} adopted separate visual–text encoders with lightweight mask decoders, but their limited language understanding restricted performance on complex queries.
Recent methods leverage VLMs for stronger textual reasoning: VISA~\cite{yan2024visa} guides SAM~\cite{kirillov2023sam1} on keyframes and propagates masks using XMem~\cite{cheng2022xmem}, while VRS-HQ~\cite{gong2025vrshq} and RGA3~\cite{wang2025rga3} build on SAM2~\cite{ravi2024sam2} with similar keyframe-based propagation schemes.
In contrast, methods such as InstructSeg~\cite{Wei_2025_instructseg} and HyperSeg~\cite{Wei_2025_hyperseg} perform frame-wise segmentation with visual-perceiver modules, without relying on mask-propagation decoders.

\paragraph{Reasoning Segmentation.}
Referring image segmentation traditionally focused on spatial grounding~\cite{yu2018mattnet,ye2019cmsa,hu2020brinet,yang2022lavt}.
LISA~\cite{lai2024lisa} introduced reasoning-aware segmentation by coupling SAM~\cite{kirillov2023sam1} with a VLM.
GLaMM~\cite{rasheed2024glamm} generalized this idea with a multi-granularity grounding design, while PixelLM~\cite{ren2024pixellm} directly mapped VLM embeddings to a mask generator through a segmentation codebook.
This paradigm was recently extended to videos via VISA~\cite{yan2024visa} and VideoLISA~\cite{bai2024videolisa_reasonvos}, with ViLLa~\cite{zheng2025villa} further improving video reasoning through enhanced context synthesis and key-segment extraction.
\section {Method}
\label{section:method}

\begin{figure*}[t]
    \setlength{\abovecaptionskip}{6pt}
    \setlength{\belowcaptionskip}{-6pt}
  \centering
    \includegraphics[width=0.9\linewidth]{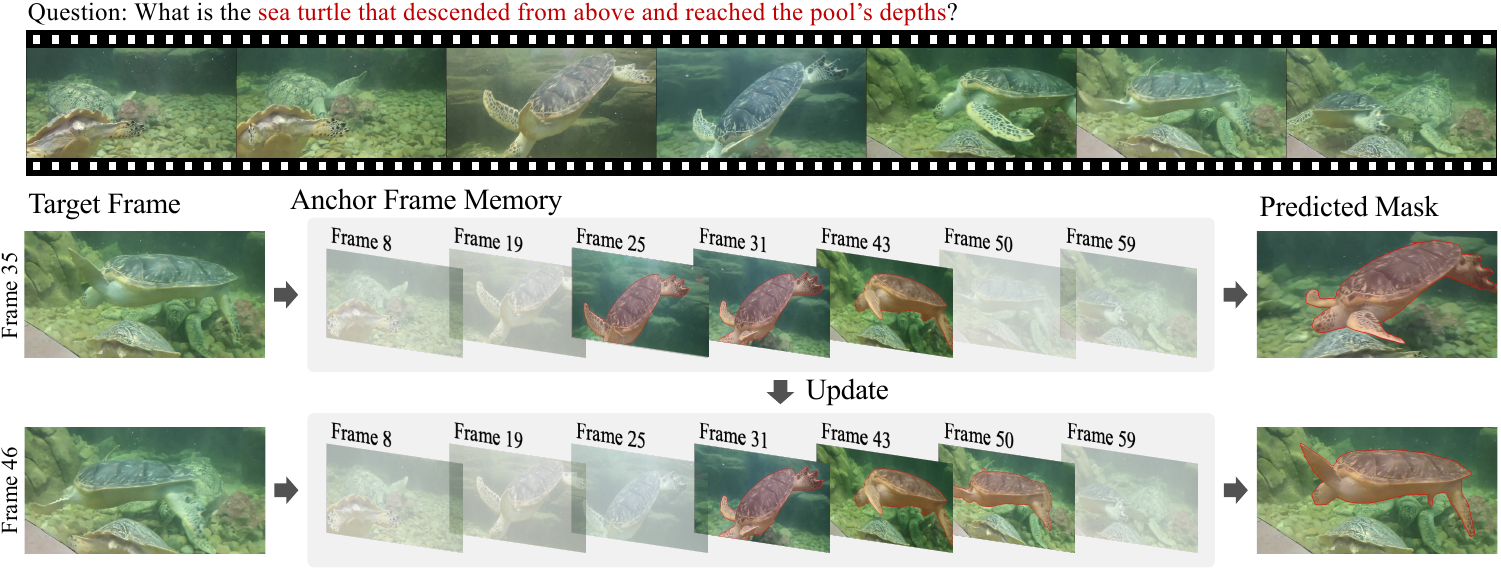} 
    \caption{
    \textbf{Anchor selection scheme of \memorymoduleabbr{}.}
    Given a video, \memorymoduleabbr{} selects \textit{anchor-frame candidates} $\mathcal{A}$ and generates their segmentation masks using frame-wise segmentation prompts. For each non-anchor frame, the module retrieves the $\alpha$ temporally nearest anchor frames $\mathcal{I}^{(k)}_{\text{Anchor}}$. The anchor set is temporally updated over time as the video advances. 
    This strategy maintains temporal locality while ensuring coverage of a broader temporal range.
    }
    \label{fig:method_anchor_selection}
\end{figure*}

\subsection{Problem Formulation}

Referring Video Object Segmentation (RVOS) aims to localize and segment objects in a video based on a natural-language expression. Given a video sequence
\begin{equation}
    \mathbf{V} = \{f_t\}_{t=1}^{T},
\end{equation}
and a referring expression $e$ that specifies a set of target objects
\begin{equation}
    \mathcal{O} = \{o_i\}_{i=1}^{N},
\end{equation}
where $f_t$ denotes the $t$-th frame of the video and $o_i$ represents the $i$-th referred object.  
Each object $o_i$ is associated with a binary segmentation mask at every frame $t$,
denoted by $\mathcal{M}_t^{o_i} \in \{0,1\}^{H \times W}$.  
The corresponding mask sequence of object $o_i$ can thus be expressed as
\begin{equation}
    \mathcal{M}^{o_i} = \{\mathcal{M}_t^{o_i}\}_{t=1}^{T}.
\end{equation}
The objective of RVOS is to identify the objects referred to by $e$ and predict their 
aggregated sequence of segmentation masks:
\begin{equation}
    \mathcal{M}_t^{\mathcal{O}} = \bigvee_{i=1}^{N} \mathcal{M}_t^{o_i}, 
    \quad t = 1, \dots, T,
\end{equation}
\begin{equation}
    \mathcal{M}^{\mathcal{O}} = \{\mathcal{M}_t^{\mathcal{O}}\}_{t=1}^{T},
\end{equation}
where $\bigvee$ denotes the element-wise (pixelwise) logical OR operation over binary masks. Following the standard practice~\citep{perazzi2016davis-vos}, we evaluate using the \jnf{} metric, 
which jointly measures region similarity (IoU) and contour accuracy between predicted and ground-truth masks.

\subsection{Overview of \virstabbr{}}
Existing RVOS methods~\cite{yan2024visa, gong2025vrshq} typically represent an entire video with a single anchor frame, which limits robustness under large motion, appearance changes, or occlusion, often causing keyframe drift and segmentation failure. Many rely on external selection models such as LLaMA-VID~\cite{li2024llama-vid} or CLIP~\cite{cherti2023openclip}, even though the most semantically informative frame may not be optimal for segmentation. Others~\cite{Wei_2025_hyperseg, Wei_2025_instructseg} perform frame-wise inference, requiring $N$ LLM forward passes for $N$ frames, leading to high computational cost and temporal inconsistency. Moreover, most multimodal segmentation models~\cite{yan2024visa, gong2025vrshq, Wei_2025_instructseg} depend on CLIP-based features that capture semantics but lack spatial precision, while recent variants~\cite{bai2024videolisa_reasonvos, Wei_2025_hyperseg} mitigate this by concatenating segmentation features at the expense of heavy computation.

To address these limitations, we propose \virstfull{}, a unified framework that integrates global spatiotemporal reasoning with fine-grained pixel segmentation in a single forward pass of a vision–language model (VLM). As illustrated in Fig.~\ref{fig:overall_arch}, 
\virstabbr{} comprises three key components: (1) {\stmodulefull} for integrating spatially precise segmentation-aware video features with semantically rich video features; (2) {\memorymodulefull} for multi-anchor-frame selection and propagation; (3) a Progressive Training Strategy that gradually aligns cross-modal reasoning, spatial fidelity, and temporal coherence.
A learnable spatiotemporal token (\texttt{[ST]}), inspired by LISA~\cite{lai2024lisa}, serves as a bridge between vision and language, unifying dense segmentation-aware video features with high-level semantic cues. The resulting fused tokens guide the mask decoder to produce temporally consistent predictions across frames.

\subsection{Spatio-Temporal Fusion (STF)}
\label{subsection:st-module}

CLIP-based vision encoders provide strong semantic representations for text–vision alignment, while the vision encoders in segmentation models capture fine-grained spatial and structural details.

To combine their complementary strengths without introducing redundant video tokens into the VLM, we design the \textbf{\stmodulefull{}}, which combines segmentation-aware video features with semantic token features, yielding compact yet expressive segmentation prompts.

Concretely, we uniformly sample $T_{\text{seg}}$ frames from the input video and feed them into the segmentation-aware vision encoder.  
The sampled RGB frames are represented as
\begin{equation}
\mathbf{V}_{\text{seg}} \in \mathbb{R}^{H \times W \times T_{\text{seg}} \times 3}.
\end{equation}
For each frame, the segmentation-aware vision encoder extracts spatial features independently, yielding
\begin{equation}
\mathbf{S}_{\text{seg}} \in \mathbb{R}^{H' \times W' \times T_{\text{seg}} \times C}.
\end{equation}
After 2D downsampling by a factor of 8, we obtain
\begin{equation}
\mathbf{S}_{\text{down}} \in \mathbb{R}^{\tfrac{H'}{8} \times \tfrac{W'}{8} \times T_{\text{seg}} \times C}.
\end{equation}
The features are then flattened into patches and projected into $D$-dimensional tokens.

We then apply \stmoduleabbr{}, which consists of two stages: \textit{Initial ST-Fusion} and \textit{Second ST-Fusion}.
In the \textit{Initial ST-Fusion} stage, the segmentation-aware video tokens $\mathbf{S}_{\text{down}}$ are first fused with learnable \texttt{[ST]} tokens $\mathbf{E}_{\text{ST}}\in \mathbb{R}^{N \times D}$ via cross-attention,
\begin{equation}
\mathbf{F}_{\text{Init}} = \mathrm{CrossAttn}\!\left(\mathbf{E}_{\text{ST}},\, \mathbf{S}_{\text{down}}\right).
\end{equation}
The fused tokens $\mathbf{F}_{\text{Init}}\in\mathbb{R}^{N\times D}$ are processed by the VLM to obtain $\mathbf{F}_{\text{ST}}\in\mathbb{R}^{N\times D}$. 
The \textit{Second ST-Fusion} stage is then applied to refine the spatiotemporal tokens. To capture temporal consistency, $\mathbf{F}_{\text{ST}}$ is temporally expanded and enriched with temporal RoPE, yielding $\mathbf{F}'_{\text{ST}}\in\mathbb{R}^{N\times T_\text{seg}\times D}$,
\begin{equation}
\mathbf{F}'_{\text{ST}} = \mathrm{Temporal Expansion}(\mathbf{F}_{\text{ST}}).
\end{equation}
Meanwhile, the segmentation-aware video tokens $\mathbf{S}_{\text{down}}$ are enriched with 3D RoPE~\cite{ma20253drope}, yielding $\mathbf{S}'_{\text{down}}$
that encode positional dynamics and enable temporal reasoning. 
We then apply cross-attention between the temporally enriched tokens $\mathbf{F}'_{\text{ST}}$ and the 3D RoPE-augmented video tokens $\mathbf{S}'_{\text{down}}$ to obtain the final spatiotemporal representation,
\begin{equation}
\tilde{\mathbf{F}}_{\text{ST}} = \mathrm{CrossAttn}\!\left(\mathbf{F}'_{\text{ST}},\, \mathbf{S}'_{\text{down}}\right)
\end{equation}
where $\tilde{\mathbf{F}}_{\text{ST}} \in \mathbb{R}^{N \times T_{\text{seg}} \times D}$. Each temporal slice corresponds to a frame-specific segmentation prompt and is applied per frame. For frame $k$, we use $\tilde{\mathbf{F}}_{\text{ST}}^{(k)} \in \mathbb{R}^{N \times 1 \times D}$.


\begin{figure*}[t]
  \centering
    \setlength{\abovecaptionskip}{6pt}
    \setlength{\belowcaptionskip}{-6pt}

    \includegraphics[width=1\linewidth]{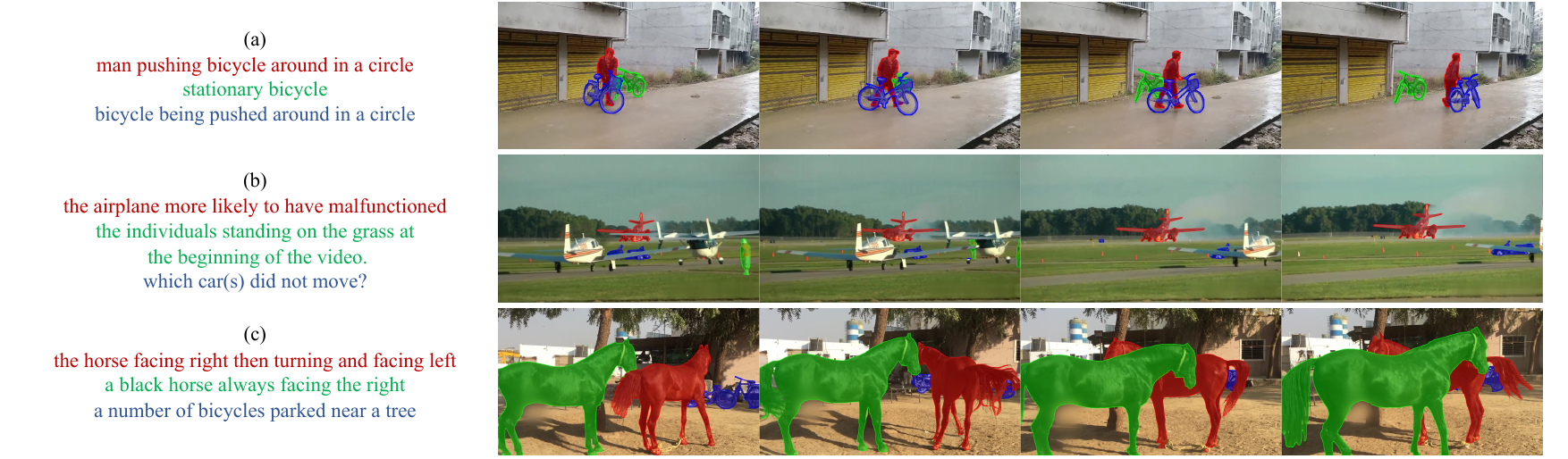} 
    \caption{
    \textbf{Qualitative results of \virstabbr{}.}
    Across diverse video segmentation scenarios, \virstabbr{} generates high-quality masks despite strong distractors, reasoning-oriented queries, heavy occlusions, small objects, and multiple interacting instances—demonstrating robust spatiotemporal reasoning and effective integration of both global and local video context. Results are best viewed when zoomed in.
    }
    \label{fig:qualitative_results}
\end{figure*}

\subsection{Temporal Dynamic Anchor Updater (TDAU)}
\label{subsection:anchor-module}

Interactive video segmentation frameworks such as SAM2~\cite{ravi2024sam2} typically rely on multiple annotated frames to initialize or refine object masks. While such multi-frame supervision (e.g., clicks or scribbles) improves stability, it is costly and infeasible at scale.

Motivated by this idea, we design an automatic mechanism that leverages the VLM’s ability to generate multiple segmentation prompts over time based solely on text descriptions.
Specifically, we introduce \textbf{\memorymodulefull{}}, which designates \textit{anchor-frame candidates} $\mathcal{A}$ and updates $\alpha$ \textit{anchor frames} selected from them over time. The \textit{anchor-frame candidates} are uniformly sampled from the $T_{\text{seg}}$ frames during inference time and randomly sampled during training (see Appendix \ref{supp:anchor-frame-selection}).

During inference, if frame $t$ belongs to the anchor frame candidate set $\mathcal{A}$, it is directly segmented using the corresponding segmentation prompt slice $\tilde{\mathbf{F}}_{\text{ST}}^{(t)}$. Otherwise, we utilize two types of memory: \textit{anchor-frame memory} and \textit{FIFO memory}. Both memory types store features $\mathbf{h}_k$ obtained by encoding the predicted mask $\hat{\mathcal{M}}_k$ together with the corresponding image features $\mathbf{S}_{\text{seg}}^{(k)}$.

For anchor-frame memory at frame $t$, we select an anchor frame index set $\mathcal{I}^{(t)}_{\text{Anchor}} \subset \mathcal{A}$ that are temporally closest to frame $t$, and use their memory features $\mathbf{h}_k$ for $k \in \mathcal{I}^{(t)}_{\text{Anchor}}$. The FIFO memory stores the features $\mathbf{h}_k$ for $k \in \mathcal{I}^{(t)}_{\text{FIFO}} = \{t-P+1, \dots, t-1\}$. 
The mask prediction at time $t$ is defined as:

\begin{equation}
\hat{\mathcal{M}}_t =
\begin{cases}
\mathcal{D}\big(\mathbf{S}_{\text{seg}}^{(t)},\, \tilde{\mathbf{F}}_{\text{ST}}^{(t)}\big), 
& t \in \mathcal{A}, \\[4pt]
\mathcal{D}\big(\mathbf{S}_{\text{seg}}^{(t)},\,
\{\mathbf{h}_k\}_{k \in \mathcal{I}^{(t)}_{\text{Anchor}}},
\{\mathbf{h}_k\}_{k \in \mathcal{I}^{(t)}_{\text{FIFO}}}\big),
& t \notin \mathcal{A}
\end{cases}
\label{eq:inference_scheme}
\end{equation}

where $\mathcal{D}$ denotes the mask decoder, $\hat{\mathcal{M}}_t$ is predicted mask at frame $t$. This formulation couples \textit{global anchor-based reasoning} with \textit{local temporal propagation}, yielding robust and consistent segmentation even under rapid motion, occlusion, and reappearance. The anchor-frame update process is illustrated in Fig.~\ref{fig:method_anchor_selection}. Notably, the anchor-frame and the FIFO memory features are fused with the current frame via cross-attention; see Appendix~\ref{supp:anchor-memory-attention} for details.

Empirically, performance improves consistently with larger $\alpha$ (see Section~\ref{subsubsec:ablation_anchor_num}), and we set $\alpha = 3$ in all main experiments for a balanced trade-off between accuracy and efficiency.

\begin{table*}[t]
  \centering
  \setlength{\tabcolsep}{4pt}
  \renewcommand{\arraystretch}{0.95}
  \footnotesize
  \caption{Performance comparison with previous methods on the ReVOS benchmark. Best results are in \textbf{bold}; 
  second-best are \underline{underlined}. 
  $\mathcal{R}$ is the robustness score for hallucination evaluations~\cite{yan2024visa, li2023r2vos}.}
  \begin{tabular}{l l l ccc ccc ccc c}
    \toprule
    \multicolumn{1}{c}{\multirow{2}{*}{\textbf{Category}}} &
    \multicolumn{1}{c}{\multirow{2}{*}{\textbf{Model}}} &
    \multicolumn{1}{c}{\multirow{2}{*}{\textbf{Venue}}} &
    \multicolumn{3}{c}{\textbf{Referring}} &
    \multicolumn{3}{c}{\textbf{Reasoning}} &
    \multicolumn{3}{c}{\textbf{Overall}} &
    \multirow{2}{*}{$\mathcal{R}$} \\
    \cmidrule(lr){4-6}\cmidrule(lr){7-9}\cmidrule(lr){10-12}
    & & &
    $\mathcal{J}$ & $\mathcal{F}$ & $\mathcal{J\&F}$ &
    $\mathcal{J}$ & $\mathcal{F}$ & $\mathcal{J\&F}$ &
    $\mathcal{J}$ & $\mathcal{F}$ & $\mathcal{J\&F}$ &
    \\
    \midrule

    \multirow{3}{*}{\parbox{2.1cm}{\centering{Segmentation\\Expert}}}
      & MTTR~\cite{botach2022mttr} & ECCV’22 & 29.8 & 30.2 & 30.0 & 20.4 & 21.5 & 21.0 & 25.1 & 25.9 & 25.5 & 5.6 \\
      & ReferFormer~\cite{wu2022referformer} & CVPR’22 & 31.2 & 34.3 & 32.7 & 21.3 & 25.6 & 23.4 & 26.2 & 29.9 & 28.1 & 8.8 \\
      & LMPM~\cite{ding2023mevis} & ICCV’23 & 29.0 & 39.1 & 34.1 & 13.3 & 24.8 & 19.0 & 21.2 & 27.1 & 26.8 & 3.8 \\
    \midrule

    \multirow{9}{*}{\parbox{2.1cm}{\centering{MLLM-based\\Segmentation\\Method}}}
      & LISA-7B~\cite{lai2024lisa} & CVPR’24 & 44.3 & 47.1 & 45.7 & 33.8 & 38.4 & 36.1 & 39.1 & 42.7 & 40.9 & 9.3 \\
      & VISA-7B~\cite{yan2024visa} & ECCV’24 & 49.2 & 52.6 & 50.9 & 40.6 & 45.4 & 43.0 & 44.9 & 49.0 & 46.9 & 15.5 \\
      & VISA-13B~\cite{yan2024visa} & ECCV’24 & 55.6 & 59.1 & 57.4 & 42.0 & 46.7 & 44.3 & 48.8 & 52.9 & 50.9 & 15.5 \\
      & HyperSeg~\cite{Wei_2025_hyperseg} & CVPR’25 & 56.0 & 60.9 & 58.5 & 50.2 & 55.8 & 53.0 & 53.1 & 58.4 & 55.7 & -- \\
      & VRS-HQ-7B~\cite{gong2025vrshq} & CVPR’25 & 59.8 & 64.5 & 62.1 & 53.5 & 58.7 & 56.1 & 56.6 & 61.6 & 59.1 & 19.7 \\
      & VRS-HQ-13B~\cite{gong2025vrshq} & CVPR’25 & \underline{61.1} & \underline{65.5} & \underline{63.3} & \underline{54.1} & \underline{59.4} & \underline{56.8} & \underline{57.6} & \underline{62.5} & \underline{60.0} & 18.9 \\
      & InstructSeg~\cite{Wei_2025_instructseg} & ICCV’25 & 54.8 & 59.2 & 57.0 & 49.2 & 54.7 & 51.9 & 52.0 & 56.9 & 54.5 & -- \\
      & RGA3-7B~\cite{wang2025rga3} & ICCV’25 & 58.7 & 62.3 & 60.5 & 53.1 & 57.7 & 55.4 & 55.9 & 60.0 & 58.0 & \textbf{28.6} \\
      & ViLLa-6B~\cite{zheng2025villa} & ICCV’25 & -- & -- & -- & -- & -- & -- & 54.9 & 59.1 & 57.0 & -- \\
    \midrule

      & \textbf{Ours (VIRST)} & -- &
        \textbf{68.8} & \textbf{72.8} & \textbf{70.8} &
        \textbf{63.9} & \textbf{68.3} & \textbf{66.1} &
        \textbf{66.3} & \textbf{70.6} & \textbf{68.4} &
        \underline{21.8} \\
    \bottomrule
  \end{tabular}
  \label{tab:comparison_revos}
\end{table*}

\begin{table*}[t]
  \centering
  \setlength{\tabcolsep}{4pt}
  \renewcommand{\arraystretch}{0.95}
  \footnotesize
  \caption{Performance comparison with previous methods on the validation sets of RVOS datasets. The best results are shown in \textbf{bold}, and the second-best results are \underline{underlined}.}
  \begin{tabular}{l l l ccc ccc ccc}
    \toprule
    \multicolumn{1}{c}{\multirow{2}{*}{\textbf{Category}}} &
    \multicolumn{1}{c}{\multirow{2}{*}{\textbf{Model}}} &
    \multicolumn{1}{c}{\multirow{2}{*}{\textbf{Venue}}} &
    \multicolumn{3}{c}{\textbf{MeViS}} &
    \multicolumn{3}{c}{\textbf{Ref-YT-VOS}} &
    \multicolumn{3}{c}{\textbf{Ref-DAVIS17}} \\
    \cmidrule(lr){4-6}\cmidrule(lr){7-9}\cmidrule(lr){10-12}
    & & &
    $\mathcal{J}$ & $\mathcal{F}$ & $\mathcal{J\&F}$ &
    $\mathcal{J}$ & $\mathcal{F}$ & $\mathcal{J\&F}$ &
    $\mathcal{J}$ & $\mathcal{F}$ & $\mathcal{J\&F}$ \\
    \midrule

    \multirow{5}{*}{\parbox{2.1cm}{\centering{Segmentation\\Expert}}}
      & ReferFormer~\cite{wu2022referformer} & CVPR’22 & 29.8 & 32.2 & 31.0 & 61.3 & 64.6 & 62.9 & 58.1 & 64.1 & 61.1 \\
      & OnlineRefer~\cite{Wu_2023_onlinerefer} & ICCV’23 & - & - & - & 61.6 & 65.5 & 63.5 & 61.6 & 67.7 & 64.8 \\
      & SAMWISE~\cite{cuttano2025samwise} & CVPR’25 & 49.5 & 46.6 & 52.4 & 69.2 & 67.8 & 70.6 & 70.6 & 67.4 & 74.5 \\
      & MPG-SAM2~\cite{rong2025mpgsam2} & ICCV’25 & 50.7 & \underline{56.7} & \underline{53.7} & \underline{71.7} & \textbf{76.1} & \underline{73.9} & 68.8 & 76.0 & 72.4 \\
      & ReferDINO~\cite{liang2025referdino} & ICCV’25 & 44.7 & 53.9 & 49.3 & 67.0 & 71.5 & 69.3 & 65.1 & 72.9 & 68.9 \\
    \midrule

    \multirow{10}{*}{\parbox{2.1cm}{\centering{MLLM-based\\Segmentation\\Method}}}
      & LISA-7B~\cite{lai2024lisa} & CVPR’24 & 35.1 & 39.4 & 37.2 & 53.4 & 54.3 & 53.9 & 62.2 & 67.3 & 64.8 \\
      & VISA-7B~\cite{yan2024visa} & ECCV’24 & 40.7 & 46.3 & 43.5 & 59.8 & 63.2 & 61.5 & 66.3 & 72.5 & 69.4 \\
      & VISA-13B~\cite{yan2024visa} & ECCV’24 & 41.8 & 47.1 & 44.5 & 61.4 & 64.7 & 63.0 & 67.0 & 73.8 & 70.4 \\
      & VideoLISA~\cite{bai2024videolisa_reasonvos} & NeurIPS’24 & 41.3 & 47.6 & 44.4 & 61.7 & 65.7 & 63.7 & 64.9 & 72.7 & 68.8 \\
      & VideoGLaMM~\cite{munasinghe2025videoglamm} & CVPR’25 & 42.1 & 48.2 & 45.2 & 65.4 & 68.2 & 66.8 & 65.6 & 73.3 & 69.5 \\
      & HyperSeg~\cite{Wei_2025_hyperseg} & CVPR’25 & - & - & - & - & - & 68.5 & - & - & 71.2 \\
      & VRS-HQ-7B~\cite{gong2025vrshq} & CVPR’25 & 47.6 & 53.7 & 50.6 & 68.3 & 72.5 & 70.4 & \underline{72.6} & \underline{79.4} & \underline{76.0} \\
      & VRS-HQ-13B~\cite{gong2025vrshq} & CVPR’25 & \underline{48.0} & 53.7 & 50.9 & 69.0 & \underline{73.1} & 71.0 & 71.0 & 77.9 & 74.4 \\
      & InstructSeg~\cite{Wei_2025_instructseg} & ICCV’25 & - & - & - & 65.4 & 69.5 & 67.5 & 67.3 & 74.9 & 71.1 \\
      & ViLLa-6B~\cite{zheng2025villa} & ICCV’25 & 46.5 & 52.3 & 49.4 & 64.6 & 70.4 & 67.5 & 70.6 & 78.0 & 74.3 \\
    \midrule
      & \textbf{Ours (VIRST)} & -- &
        \textbf{60.4} & \textbf{65.4} & \textbf{62.9} &
        \textbf{72.2} & \textbf{76.1} & \textbf{74.2} &
        \textbf{75.9} & \textbf{83.1} & \textbf{79.5} \\
    \bottomrule
  \end{tabular}
  \label{tab:comparison_rvos}
\end{table*}

\begin{table*}[t]
  \centering
  
    \setlength{\abovecaptionskip}{6pt}
    \setlength{\belowcaptionskip}{-6pt}
  \setlength{\tabcolsep}{5pt}
  \renewcommand{\arraystretch}{0.95}
  \footnotesize
  \caption{Performance comparison with previous MLLM-based segmentation methods on referring and reasoning image segmentation datasets. The best results are shown in \textbf{bold}, and the second-best results are \underline{underlined}.}
  \begin{tabular}{lcccccccccc}
    \toprule
    \multirow{2}{*}{\textbf{Model}} 
      & \multicolumn{3}{c}{RefCOCO} 
      & \multicolumn{3}{c}{RefCOCO+} 
      & \multicolumn{2}{c}{RefCOCOg} 
      & \multicolumn{2}{c}{ReasonSeg} \\
    \cmidrule(lr){2-4} \cmidrule(lr){5-7} \cmidrule(lr){8-9} \cmidrule(lr){10-11}
      & val & testA & testB
      & val & testA & testB
      & val(U) & test(U)
      & gIoU & cIoU \\
    \midrule
    LISA-7B~\citep{lai2024lisa} & 74.9 & 79.1 & 72.3 & 65.1 & 70.8 & 58.1 & 67.9 & 70.6 & 52.9 & 54.0 \\
    PixelLM-7B~\citep{ren2024pixellm} & 73.0 & 76.5 & 68.2 & 66.3 & 71.7 & 58.3 & 69.3 & 70.5 & - & - \\
    GLaMM~\citep{rasheed2024glamm} & 79.5 & 83.2 & 76.9 & 72.6 & 78.7 & 64.6 & 74.2 & 74.9 & - & - \\
    VISA-7B~\citep{yan2024visa} & 72.4 & 75.5 & 68.1 & 59.8 & 64.8 & 53.1 & 65.5 & 66.4 & 52.7 & 57.8 \\
    VRS-HQ-7B~\citep{gong2025vrshq} & 73.5 & 77.5 & 69.5 & 61.7 & 67.6 & 64.3 & 66.7 & 67.5 & 51.7 & 52.9 \\ 
    HyperSeg~\citep{Wei_2025_hyperseg} & 84.8 & 85.7 & \underline{83.4} & \underline{79.0} & 83.5 & \underline{75.2} & \underline{79.4} & 78.9 & 59.2 & 56.7 \\
    InstructSeg~\citep{Wei_2025_instructseg} & \underline{85.8} & \underline{86.6} & \textbf{84.0} & \textbf{80.1} & \underline{83.8} & \textbf{75.6} & 79.3 & \underline{80.3} & \textbf{61.9} & \textbf{65.2} \\ 
    \midrule
    \textbf{Ours (VIRST)} & \textbf{86.4} & \textbf{90.7} & 81.2 & 78.2 & \textbf{84.6} & 68.8 & \textbf{82.1} & \textbf{83.1} & \underline{60.8} & \underline{65.0} \\
    \bottomrule
  \end{tabular}
  \label{tab:comparison_refcoco_reaseg}
\end{table*}

\subsection{Progressive Training Strategy}
\label{subsection:training_strategy}
Directly fine-tuning the entire network from scratch often results in unstable optimization and poor convergence due to the newly introduced modules and long-range temporal dependencies. To stabilize training and enable gradual adaptation of the vision–language backbone, we employ a {three-stage progressive curriculum} that aligns semantics first and temporal reasoning later.

\paragraph{(1) Alignment Stage.}  
We first freeze the mask prediction and memory modules of SAM2~\cite{ravi2024sam2}, and train only the \stmoduleabbr\ together with the LoRA adapters inserted into the VLM backbone. This stage focuses on aligning the fused spatio-temporal features with the language-conditioned reasoning capability of the backbone while keeping the base segmentation components stable.
We adopt an \emph{image-like training} paradigm to provide a stable initialization for temporal learning.  
We randomly sample an anchor frame index set $\mathcal{A}_{\text{train}}$ with $|\mathcal{A}_{\text{train}}|=\alpha$ from each video and predict segmentation masks independently without propagation, formulated as:
\begin{equation}
\begin{aligned}
\mathcal{A}_{\text{train}} 
&\sim \mathrm{Uniform}\big(\{1,\dots,T_{\text{seg}}\}\big)\ \text{without replacement} \\[4pt]
\hat{\mathcal{M}}_{t} &= 
\mathcal{D}\!\left(\mathbf{S}^{(t)}_{\text{seg}},\, \tilde{\mathbf{F}}_{\text{ST}}^{(t)}\right),
\quad t \in \mathcal{A}_{\text{train}}
\end{aligned}
\label{eq:random_anchor_sampling}
\end{equation}
This step focuses on learning reliable per-frame grounding before temporal propagation is introduced.

\paragraph{(2) Few-Image Prediction Stage.}  
Next, we gradually unfreeze the VLM projection layer, the \stmoduleabbr, and the mask decoder, while still freezing the VLM video encoder, the main transformer layers of the VLM, and the segmentation-aware vision encoder. 
The training setup follows the same image-like paradigm as in Stage~1, except that more modules are unfrozen for joint optimization.

\paragraph{(3) Propagation Stage.}  
Finally, we extend training from per-frame grounding to anchor-based temporal propagation. As backpropagating gradients from all video frames is prohibitively expensive, we adopt a memory-efficient strategy that covers diverse anchor selection scenarios via random sampling. We sample an anchor index set $\mathcal{A}_{\text{train}}$ per video, as in Stages~1 and~2, and select propagation frames $\mathcal{I}_{\text{prop}}$ around the anchors using a rule-based scheme, yielding 10--20 targets per clip. Anchor frames are segmented from the spatiotemporal prompts, while propagation frames are inferred by conditioning on the anchor and FIFO memory:

\begin{equation}
\hat{\mathcal{M}}_t =
\begin{cases}
\mathcal{D}\big(\mathbf{S}_{\text{seg}}^{(t)},\, \tilde{\mathbf{F}}_{\text{ST}}^{(t)}\big), 
& t \in \mathcal{A}_{\text{train}}, \\[4pt]
\mathcal{D}\big(\mathbf{S}_{\text{seg}}^{(t)},\,
\{\mathbf{h}_{k}\}_{k \in \mathcal{A}_{\text{train}}},\,
\{\mathbf{h}_{k}\}_{k \in \mathcal{I}^{(t)}_{\text{FIFO}}}\big),
& t \in \mathcal{I}_{\text{prop}}
\end{cases}
\label{eq:training_scheme}
\end{equation}

This final stage enables VIRST to learn temporal propagation and handle occlusions while maintaining the same structure as the inference scheme (Eq.~\ref{eq:inference_scheme}). Details of the anchor-frame selection mechanism are provided in Appendix~\ref{supp:anchor-frame-selection}, with additional training details in Appendix~\ref{supp:training_stages}.



\subsection{Training Objectives}

To jointly optimize spatial accuracy, cross-modal reasoning, and temporal consistency, 
we adopt a composite objective combining binary cross-entropy (BCE), Dice, token cross-entropy, occlusion, and IoU losses. 
BCE and Dice ensure pixel- and region-level segmentation fidelity, 
while the token loss $\mathcal{L}_{token}$ aligns text-conditioned reasoning within the VLM. 
The occlusion loss $\mathcal{L}_{occ}$ models object visibility, and the IoU loss $\mathcal{L}_{iou}$ provides confidence-aware temporal regularization. 
The overall objective is formulated as:

\begin{equation}
\begin{aligned}
\mathcal{L}_{total} = {} &
\lambda_{bce} \mathcal{L}_{bce} +
\lambda_{dice} \mathcal{L}_{dice} +
\lambda_{token} \mathcal{L}_{token} \\
& +
\lambda_{occ} \mathcal{L}_{occ} +
\lambda_{iou} \mathcal{L}_{iou}.
\end{aligned}
\end{equation}

Further details of the loss formulation and training setup are provided in Appendix~\ref{supp:training}.
\section{Experiments}
\subsection{Experimental Settings}

\begin{figure}[t]
  \centering
    \setlength{\abovecaptionskip}{6pt}
    \setlength{\belowcaptionskip}{-14pt}
    \includegraphics[width=1\linewidth]{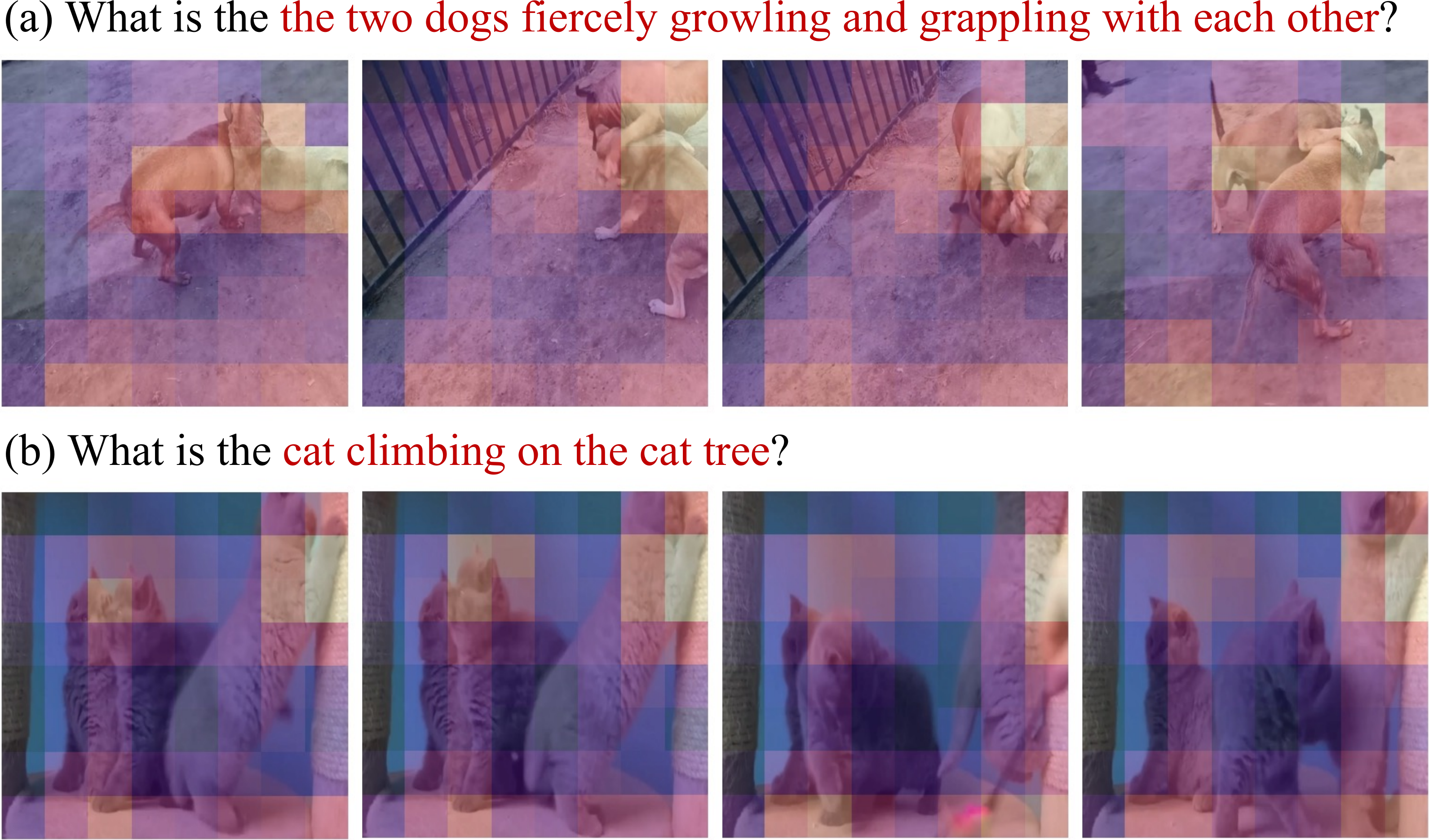} 
    \caption{
    \textbf{\stmoduleabbr{} patch-wise attention visualization.}
    We visualize the $8 \times 8$ patch-level attention from the \stmoduleabbr{} before feeding it into the segmentation decoder. The attention maps consistently highlight key motion regions along the spatiotemporal dimension.
    }
    \label{fig:st-attention}
\end{figure}

In our experiments, the vision–language backbone was initialized with VideoChat-Flash-7B~\cite{li2024videochatflash}, a VLM whose vision encoder is a ViT-based model pretrained with UMT~\cite{vit, li2023umt}, while the mask prediction branch adopted the SAM2 architecture~\cite{ravi2024sam2}. Low-rank adaptation (LoRA)~\cite{hu2022lora} was applied to the VLM for efficient fine-tuning, while the proposed \stmodulefull{} module was trained from scratch.

We trained the model on a comprehensive set of datasets to ensure both temporal reasoning and segmentation generalization. The training corpus includes four Referring Video Object Segmentation (RVOS) datasets, namely Ref-DAVIS17~\cite{khoreva2018ref-davis17}, Ref-YouTube-VOS~\cite{seo2020urvos}, MeViS~\cite{ding2023mevis}, and ReVOS~\cite{yan2024visa}, as well as the video instance segmentation dataset LV-VIS~\cite{wang2023lvvis}. To encourage robust multimodal grounding, we further incorporated referring image segmentation datasets (RefCOCO, RefCOCO+, and RefCOCOg~\cite{kazemzadeh-etal-2014-referitgame}), semantic segmentation datasets (ADE20k~\cite{ade20k}, COCO-Stuff~\cite{caesar2018cocostuff}, PACO~\cite{ramanathan2023paco}, PASCAL-Part~\cite{chen2014detect}), and image reasoning data from ReasonSeg~\cite{lai2024lisa}. Additionally, video instruction–tuning data from VideoLLaVA-Instruct~\cite{lin2024videollava} was used to strengthen text–video alignment. Training was conducted on 8 NVIDIA H100 GPUs for three days. Further implementation details are provided in Appendix~\ref{supp:exp-setup}.

\subsection{Referring Video Object Segmentation}

We first evaluate \virstabbr{} on the four representative RVOS benchmarks: ReVOS, MeViS, Ref-YT-VOS, and Ref-DAVIS17. Across all settings, \virstabbr{} achieves consistent state-of-the-art performance, outperforming prior models by significant margins.

On ReVOS (Table~\ref{tab:comparison_revos}), which contains reasoning queries that require higher-level reasoning ability beyond standard referring questions, \virstabbr{} achieves the highest performance across both \textit{referring} and \textit{reasoning} settings, surpassing the previous state-of-the-art by more than 8 points in overall \jnf. While performing strongly in both settings, the improvement is more pronounced in the reasoning case, where \virstabbr\ outperforms the previous best result by 9 points, compared to a 7.5-point margin in the referring case. This demonstrates that our method not only excels at precise visual grounding but also generalizes effectively to reasoning-oriented queries, reflecting the robustness and reliability of its overall performance.

We further evaluate our model on MeViS (valid split), Ref-YT-VOS (valid split), and Ref-DAVIS17 (valid split), as shown in Table~\ref{tab:comparison_rvos}. MeViS, which includes motion expressions describing dynamic scenes, poses a more challenging setting. On this dataset, \virstabbr{} achieves the best performance with a 9-point improvement in overall \jnf, demonstrating strong capability in modeling complex spatiotemporal dynamics. In Ref-YT-VOS and Ref-DAVIS17, it also reaches state-of-the-art performance, highlighting the model’s consistent generalization and effectiveness across diverse video scenarios. Qualitative examples in Fig.~\ref{fig:qualitative_results} and Appendix~\ref{supp:video_qualitative} illustrate the precision and robustness of the segmentation outputs.

\begin{table}[t]
  \centering
  \setlength{\tabcolsep}{6pt}
  \renewcommand{\arraystretch}{0.95}
  \footnotesize
  \caption{
    Ablation study of the Initial and Second ST-Fusion modules on the MeViS validation set. 
    \checkmark~indicates that the corresponding module is enabled.
  }
  \begin{tabular}{ccccc}
    \toprule
    \multirow{2}{*}{\textbf{Initial ST-Fusion}} & 
    \multirow{2}{*}{\textbf{Second ST-Fusion}} & 
    \multicolumn{3}{c}{\textbf{MeViS}} \\
    \cmidrule(lr){3-5}
     & & $\mathcal{J}$ & $\mathcal{F}$ & $\mathcal{J\&F}$ \\
    \midrule
    \checkmark & \checkmark & \textbf{60.4} & \textbf{65.4} & \textbf{62.9} \\
    \checkmark & & 57.1 & 62.4 & 59.7 \\
     &  & 57.0 & 61.8 & 59.4 \\
    \bottomrule
  \end{tabular}
  \label{tab:ablation_initial_st}
\end{table}

\begin{table}[t]
  \centering
  
    \setlength{\abovecaptionskip}{6pt}
    \setlength{\belowcaptionskip}{-6pt}
  \setlength{\tabcolsep}{6pt}
  \renewcommand{\arraystretch}{0.95}
  \footnotesize
  \caption{
    Ablation study of anchor frame selection strategies on the MeViS validation set.
  }
  \begin{tabular}{lccc}
    \toprule
    \multirow{2}{*}{\textbf{Anchor Frame Selection Strategy}} &
    \multicolumn{3}{c}{\textbf{MeViS}} \\
    \cmidrule(lr){2-4}
     & $\mathcal{J}$ & $\mathcal{F}$ & $\mathcal{J\&F}$ \\
    \midrule
    Dynamic anchor selection & \textbf{60.4} & \textbf{65.4} & \textbf{62.9} \\
    First-frame baseline & 55.4 & 60.4 & 57.9 \\
    CLIP-guided selection & 56.8 & 61.8 & 59.3 \\
    Random-3 sampling & 59.6 & 64.8 & 62.2 \\
    Uniform-3 sampling & 59.4 & 64.3 & 61.9 \\
    \bottomrule
  \end{tabular}
  \label{tab:anchor_selection_ablation}
\end{table}

\begin{table}[t]
  \centering
  
    \setlength{\abovecaptionskip}{6pt}
    \setlength{\belowcaptionskip}{-6pt}
  \setlength{\tabcolsep}{4pt}
  \renewcommand{\arraystretch}{0.95}
  \footnotesize
  \caption{Ablation study on the number of anchor frames ($\alpha$) used in \virstabbr{} on the MeViS validation set.}
  \begin{tabular}{cccc}
    \toprule
    \multirow{2}{*}{\makecell{\textbf{Number of Anchor Frames ($\alpha$)}}} &
    \multicolumn{3}{c}{\textbf{MeViS}} \\
    \cmidrule(lr){2-4}
     & $\mathcal{J}$ & $\mathcal{F}$ & $\mathcal{J\&F}$ \\
    \midrule
    2  & 60.1 & 65.0 & 62.5 \\
    3  & 60.4 & 65.4 & 62.9 \\
    4  & 60.4 & 65.4 & 62.9 \\
    6  & 60.6 & 65.5 & 63.0 \\
    8  & \textbf{60.7} & \textbf{65.8} & \textbf{63.2} \\
    \bottomrule
  \end{tabular}
  \label{tab:ablation_num_anchor_frames}

\end{table}

\subsection{Image Segmentation}
We examine whether the proposed design generalizes beyond videos by evaluating on RefCOCO, RefCOCO+, RefCOCOg~\citep{kazemzadeh-etal-2014-referitgame}, and the reasoning-based image segmentation dataset ReasonSeg~\citep{lai2024lisa}. As shown in Table~\ref{tab:comparison_refcoco_reaseg}, VIRST achieves the best performance on RefCOCO (val, testA), RefCOCO+ (testA), and RefCOCOg (val, test), while maintaining competitive results on the other benchmarks. These findings demonstrate that our spatiotemporal fusion mechanism generalizes strongly to image segmentation tasks. Additional qualitative results are provided in Appendix~\ref{supp:image_qualitative}.

\subsection{Ablation Studies}

\paragraph{Effect of the \stmoduleabbr{}.}
In Table~\ref{tab:ablation_initial_st}, we present an ablation study of the \stmoduleabbr{} module. We evaluate its components by selectively disabling each part: removing Initial ST-Fusion makes the \sttoken{} a solely learnable embedding, delaying integration after the VLM, and replacing the Second ST-Fusion with an MLP removes spatiotemporal structure. The results show that enabling both components yields a 3.5-point gain in \jnf, demonstrating their effectiveness in enhancing spatiotemporal reasoning and feature alignment. Figure~\ref{fig:st-attention} visualizes the attention patterns, which reveals that the fused tokens capture structural and motion-centric features that the VLM alone cannot model.

\paragraph{Effect of the \memorymoduleabbr{}.}
\label{subsubsec:ablation_anchor_num}
In Table~\ref{tab:anchor_selection_ablation}, we compare different anchor-frame selection strategies. We consider \textit{first-frame} and \textit{CLIP-guided selection} schemes, following conventional one-shot video segmentation and RVOS methods, both of which result in performance drops on MeViS. These settings correspond to removing TDAU. To further analyze the role of anchor frames, we study degenerate variants of \memorymoduleabbr{}, where we fix $\alpha=3$ and examine \textit{random-3} and \textit{uniform-3} sampling schemes, in which three anchor frames are globally selected and kept constant throughout the video. Although these fixed schemes outperform single-frame baselines, they still fall significantly behind our method, which updates anchor frames over time.

\paragraph{Effect of the number of anchor frames.}
In Table~\ref{tab:ablation_num_anchor_frames}, we analyze how performance varies with the number of anchor frames ($\alpha$). Increasing $\alpha$ consistently improves results by enabling the model to cover a broader temporal span and capture the target object’s appearance and motion, leading to more stable segmentation. The results also indicate that $\alpha$ acts as an \textit{inference-time scaling} parameter: when higher reliability is required, allocating more compute by increasing $\alpha$ yields more robust predictions. The efficiency of \virstabbr{} for different $\alpha$ is provided in Appendix~\ref{supp:efficiency}.
\section{Conclusion}

In this paper, we presented \virstfull{}, an end-to-end framework for Referring Video Object Segmentation (RVOS) that unifies global video reasoning and fine-grained mask prediction within a single model. The proposed \textbf{\stmodulefull} effectively bridges semantic and segmentation representations, while the \textbf{\memorymodulefull} enables temporal anchor-frame updates. Through this unified design, \virstabbr{} achieves strong spatiotemporal consistency and sets new state-of-the-art performance across multiple RVOS benchmarks. 
\section*{Acknowledgements}
This work was supported in part by National Research Foundation of Korea (NRF) grant (RS-2025-00560762), and Institute of Information \& communications Technology Planning \& Evaluation (IITP) grant (RS-2024-00454666, RS-2025-25442338, RS-2024-00397085, RS-2021-II211343). This research was also conducted as part of the Sovereign AI Foundation Model Project (Data Track, 2026-AIData-WII01), organized by the Ministry of Science and ICT (MSIT) and supported by the National Information Society Agency (NIA). J. Do is with ASRI, Seoul National University.
{
    \small
    \bibliographystyle{ieeenat_fullname}
    \bibliography{main}
}

\clearpage
\appendix
\setcounter{page}{1}
\maketitlesupplementary

\section{Implementation Details}

\subsection{Experimental Setup}
\label{supp:exp-setup}

\paragraph{Hardware specifications.} All experiments were run on 8$\times$ NVIDIA H100~80GB GPUs, an Intel Xeon Platinum~8480+ CPU, and 2~TB RAM.

\paragraph{Dataset compositions.} The overall dataset composition follows VISA~\cite{yan2024visa} and LISA~\cite{lai2024lisa}. The datasets used for training are listed in Table~\ref{tab:training_datasets}. We reuse the official VISA dataloader implementations with a few minor adjustments.

\begin{table}[h!]
\centering
\small
\caption{Datasets used for training VIRST.}
\label{tab:training_datasets}
\begin{tabular}{l l}
\toprule
\textbf{Category} & \textbf{Dataset} \\
\midrule

\multirow{4}{*}{RVOS} 
& Ref-DAVIS17~\cite{khoreva2018ref-davis17} \\
& Ref-YouTube-VOS~\cite{seo2020urvos} \\
& MeViS~\cite{ding2023mevis} \\
& ReVOS~\cite{yan2024visa} \\

\midrule

Video Instance Segmentation 
& LV-VIS~\cite{wang2023lvvis} \\

\midrule

\multirow{3}{*}{Referring Image Segmentation} 
& RefCOCO~\cite{kazemzadeh-etal-2014-referitgame} \\
& RefCOCO+~\cite{kazemzadeh-etal-2014-referitgame} \\
& RefCOCOg~\cite{kazemzadeh-etal-2014-referitgame} \\

\midrule

\multirow{4}{*}{Semantic Segmentation} 
& ADE20k~\cite{ade20k} \\
& COCO-Stuff~\cite{caesar2018cocostuff} \\
& PACO~\cite{ramanathan2023paco} \\
& PASCAL-Part~\cite{chen2014detect} \\

\midrule

Image Reasoning 
& ReasonSeg~\cite{lai2024lisa} \\

Video Instruction Tuning 
& VideoLLaVA-Instruct~\cite{lin2024videollava} \\

\bottomrule
\end{tabular}
\end{table}

For PACO and PASCAL-Part, we modify the loader to produce complete part-level masks when a textual reference corresponds to multiple object parts, ensuring consistent supervision across part-segmentation datasets.
In addition, a small number of ReVOS training samples contained misaligned annotations where the ground-truth mask did not match the associated frame; these corrupted instances were excluded.
All validation and test sets were used exactly as provided, and no dataset-specific modifications were applied during inference.

Additionally, a subset of the VideoLLaVA-Instruct dataset was incorporated to preserve the model’s video understanding capabilities. 
This additional supervision preserves robust semantic video understanding in \virstabbr{}, enabling it to generate natural-language responses in an autoregressive manner.
Relevant results and analysis are provided in Appendix~\ref{supp:video_understanding}.

\subsection{Training Procedure}
\label{supp:training}

\paragraph{Initialization.}
We adopt VideoChat-Flash~\cite{li2024videochatflash} as our vision–language backbone. Specifically, we use the publicly released \texttt{VideoChat-Flash-Qwen2-7B\_res448} checkpoint from HuggingFace, chosen for its strong video understanding capability and reliable reproducibility.
For the segmentation module, we employ SAM~2.1 initialized from the \texttt{sam2.1\_hiera\_large} checkpoint. All other components not explicitly mentioned are initialized from scratch.

\paragraph{Training objective.}
The overall training objective is expressed as:
\begin{equation}
\begin{split}
\mathcal{L}_{\text{total}} = {} &
\lambda_{\text{bce}}\mathcal{L}_{\text{bce}}
+ \lambda_{\text{dice}}\mathcal{L}_{\text{dice}}
+ \lambda_{\text{token}}\mathcal{L}_{\text{token}} \\
& + \lambda_{\text{occ}}\mathcal{L}_{\text{occ}}
+ \lambda_{\text{iou}}\mathcal{L}_{\text{iou}}.
\end{split}
\end{equation}

In all experiments, we set 
$\lambda_{\text{bce}} = 1.0$, 
$\lambda_{\text{dice}} = 1.0$, 
$\lambda_{\text{token}} = 1.0$, 
$\lambda_{\text{occ}} = 0.05$, 
and $\lambda_{\text{iou}} = 0.05$,  
to balance segmentation fidelity, reasoning alignment, and temporal smoothness.

\paragraph{Training details.}
We train the model using the AdamW optimizer. Training is performed in \texttt{bfloat16} with a linear warmup of $100$ steps followed by a decay over the full schedule. We adopt ZeRO stage~2, a per-GPU micro-batch size of $1$ (due to memory constraints from high-resolution video inputs), and $16$-step gradient accumulation. All segmentation supervision is applied at a resolution of $1024\times1024$.

\paragraph{Dataset ratio.}
We group the training data into five categories: semantic segmentation, referring image segmentation, reasoning-based image segmentation, RVOS and video VQA. During training, samples are drawn with category-wise sampling weights of $[4,\,3,\,1,\,12,\,1]$.

\subsection{Training Stages}
\label{supp:training_stages}

\paragraph{\stageone. } 
We freeze all modules except the \stmoduleabbr, the LM head, and the LoRA adapters.
A constant learning rate of $2\times10^{-4}$ is used.

\paragraph{\stagetwo.}
We continue with the same learning rate of $2\times10^{-4}$ and unfreeze the mask decoder, memory attention module, memory encoder, and the multi-modal projector. 
This stage enables full segmentation capability while keeping the VLM backbone mostly stable.

\paragraph{\stagethree.}
Same freezing configuration as in Stage~2, but the learning rate is reduced to $1\times10^{-5}$. 
This stage additionally activates propagation-based supervision to refine temporal consistency.

\paragraph{Results after each training stage.}
Table~\ref{tab:stage_results} summarizes performance after Stages~1–3 on the MeViS valid\_u split. 
Each stage provides consistent improvements, with the final stage achieving the highest $\mathcal{J}$, $\mathcal{F}$, and $\mathcal{J\&F}$ scores.

\begin{table}[h!]
  \centering
  \renewcommand{\arraystretch}{0.95}
  \setlength{\tabcolsep}{10pt}
  \footnotesize
  \caption{
    Performance after each training stage on MeViS (valid\_u).
  }
  \label{tab:stage_results}
  \begin{tabular}{cccc}
    \toprule
    \multirow{2}{*}{\textbf{Stage}} &
    \multicolumn{3}{c}{\textbf{MeViS (valid\_u)}} \\
    \cmidrule(lr){2-4}
     & $\mathcal{J}$ & $\mathcal{F}$ & $\mathcal{J\&F}$ \\
    \midrule
    Stage\,1 & 57.6 & 63.6 & 60.6 \\
    Stage\,2 & 61.1 & 67.8 & 64.4 \\
    Stage\,3 & 69.6 & 75.7 & 72.6 \\
    \bottomrule
  \end{tabular}
\end{table}

\paragraph{Training process ablation study.}
\begin{table}[h!]
  \centering
  \renewcommand{\arraystretch}{0.95}
  \footnotesize
  \caption{
    Ablation on training stage combinations on MeViS (valid\_u).
  }
  \label{tab:supp_stage_results_ablation}
  \begin{tabular}{cccc}
    \toprule
    \multirow{2}{*}{\textbf{Training Stages}} &
    \multicolumn{3}{c}{\textbf{MeViS (valid\_u)}} \\
    \cmidrule(lr){2-4}
     & $\mathcal{J}$ & $\mathcal{F}$ & $\mathcal{J\&F}$ \\
    \midrule
    Full Pipeline (Stage\,1+2+3) & \textbf{69.6} & \textbf{75.7} & \textbf{72.6} \\
    Without Stage\,1 (Stage\,2+3)   & 62.8 & 68.8 & 65.8 \\
    Stage\,3 Only & 62.4 & 69.1 & 65.8 \\
    \bottomrule
  \end{tabular}
\end{table}

Table~\ref{tab:supp_stage_results_ablation} presents the ablation study on different training-stage configurations.
For the first variant, we removed \stageone{} (Stage 1) by replacing it with the same configuration as \stagetwo{} (Stage 2), unfreezing all modules from the beginning.
For the \stagethree{} (Stage~3) only setting, both \stageone{} and \stagetwo{} were replaced with the \stagethree{} configuration.
Across all configurations, the total number of epochs and all hyperparameters were kept identical to ensure fair comparison.
As shown in the table, incorporating all three stages described in Section~\ref{subsection:training_strategy} is crucial for achieving strong performance.

\section{Architectural Details}
\label{supp:architectural-details}

\subsection{Anchor Frame Selection}
\label{supp:anchor-frame-selection}

\subsubsection{Training}






\begin{algorithm}[h]
\small
\caption{Anchor-Frame and Propagation-Frame Sampling during Training}
\label{alg:random_keyframe_selection}
\SetKwInOut{Input}{Input}\SetKwInOut{Output}{Output}

\Input{Video length $T_{\text{seg}}$, maximum propagation window $n_{\text{prop}}$}
\Output{Anchor-frame index set $\mathcal{A}_{\text{train}}$, propagation-frame index set $\mathcal{I}_{\text{prop}}$}

Select $\mathcal{A}_{\text{train}} \leftarrow$ randomly sample $\min(\alpha, T_{\text{seg}})$ distinct frame indices from $\{0, \ldots, T_{\text{seg}}-1\}$\;
Sort $\mathcal{A}_{\text{train}}$ in ascending order\;

Initialize $\mathcal{I}_{\text{prop}} \leftarrow \emptyset$\;

\ForEach{$k \in \mathcal{A}_{\text{train}}$}{
    Add preceding frame indices $\{k-2, k-1\}$ that lie within range to $\mathcal{I}_{\text{prop}}$\;
    Add succeeding frame indices $\{k+1, k+2, \ldots, k+n_{\text{prop}}\}$ that lie within range to $\mathcal{I}_{\text{prop}}$\;
}
Remove any elements of $\mathcal{A}_{\text{train}}$ from $\mathcal{I}_{\text{prop}}$\;
Sort $\mathcal{I}_{\text{prop}}$ in ascending order\;

\Return $\mathcal{A}_{\text{train}}, \mathcal{I}_{\text{prop}}$
\end{algorithm}

Alg.~\ref{alg:random_keyframe_selection} outlines the training-time anchor-frame selection strategy. We randomly sample up to $\alpha$ anchor frames $\mathcal{A}_{\text{train}}$ from a video and collect their local temporal neighbors as propagation frames, enabling the model to learn propagation cues while keeping memory usage feasible for high-resolution mask prediction.

Specifically, for each anchor frame, we include two preceding frames and up to $n_{\text{prop}}$ subsequent frames as propagation targets. In our setting, $\alpha=3$ throughout training and $n_{\text{prop}}=5$.

\subsubsection{Inference}
\begin{algorithm}[h]
\small
\caption{Anchor-Frame Selection and Update during Inference}
\label{alg:inference_anchor_selection}
\SetKwInOut{Input}{Input}\SetKwInOut{Output}{Output}

\Input{Video length $T_{\text{seg}}$, number of selected anchors $\alpha$}
\Output{Anchor-frame index set $\mathcal{A}$, per-frame anchor subset $\mathcal{I}^{(t)}_{\text{Anchor}}$}

Set $K \leftarrow \max(1, \lfloor T_{\text{seg}} / 4 \rfloor)$\;

Uniformly sample $K$ anchor indices from $\{0, \ldots, T_{\text{seg}}-1\}$ to form $\mathcal{A}$\;
Sort $\mathcal{A}$ in ascending order\;

\For{$t = 0$ \KwTo $T_{\text{seg}}-1$}{
    Compute distances $d(k, t) = |k - t|$ for all $k \in \mathcal{A}$\;
    
    Sort $\mathcal{A}$ by increasing $d(k,t)$\;
    
    Select $\mathcal{I}^{(t)}_{\text{Anchor}} \leftarrow$ first $\min(\alpha, |\mathcal{A}|)$ elements\;
}

\Return $\mathcal{A}, \{\mathcal{I}^{(t)}_{\text{Anchor}}\}_{t=0}^{T_{\text{seg}}-1}$
\end{algorithm}
At inference, we uniformly sample an anchor frame set $\mathcal{A}$ from the $T_{\text{seg}}$ frames. We set $|\mathcal{A}| = \max(1, \lfloor T_{\text{seg}} / 4 \rfloor)$. Since $T_{\text{seg}}$ is capped at 32, this yields at most 8 anchor frames for longer videos, corresponding to a stride of $\Delta T_{\text{seg}} = 4$. 

For mask prediction at time $t$, we select the $\alpha$ anchor frames closest to $t$ from $\mathcal{A}$ to form the set $\mathcal{I}^{(t)}_{\text{Anchor}}$, as detailed in Alg.~\ref{alg:inference_anchor_selection}.

\subsection{Anchor-Frame Memory Attention}
\label{supp:anchor-memory-attention}
For each frame $t$, we construct a unified memory token sequence from the anchor frame index set $\mathcal{I}^{(t)}_{\text{Anchor}}$ and the FIFO index set $\mathcal{I}^{(t)}_{\text{FIFO}}$, which contains the indices of the most recent $P$ frames. The anchor-frame memory captures long-range context, while the FIFO memory focuses on recent frames.

For anchor-frame memory attention, we assign a temporal index
\begin{equation}
\tau(k) =
\begin{cases}
0, & k \in \mathcal{I}^{(t)}_{\text{Anchor}},\\
1,2,\dots,P, & k \in \mathcal{I}^{(t)}_{\text{FIFO}},
\end{cases}
\end{equation}
which modulates a learned temporal positional encoding $\mathrm{PE}(\tau(k))$.

The memory tokens are constructed as
\begin{equation}
\mathbf{H}_t =
\big[\, \mathbf{h}_k + \mathrm{PE}(\tau(k)) \,\big]_{k \in \mathcal{I}^{(t)}_{\text{Anchor}} \cup \mathcal{I}^{(t)}_{\text{FIFO}}},
\end{equation}

where $[\cdot]$ denotes concatenation along the token dimension.

Given the current-frame features $\mathbf{S}^{(t)}_{\text{seg}}$, anchor memory attention produces memory-conditioned features
\begin{equation}
\tilde{\mathbf{S}}_{\text{seg}}^{(t)} =
\mathrm{CrossAttn}\!\left(
\mathbf{S}^{(t)}_{\text{seg}},\; \mathbf{H}_t
\right),
\label{eq:ama_final_revised}
\end{equation}
where $\tau(k)=0$ encodes invariant anchor context and $\tau(k)>0$ captures recency-aware FIFO cues.

\subsection{Frame-Aware Video Tokenizer}

\label{supp:video_tokenizer}
\begin{figure}[h!]
  \centering
    \includegraphics[width=1\linewidth]{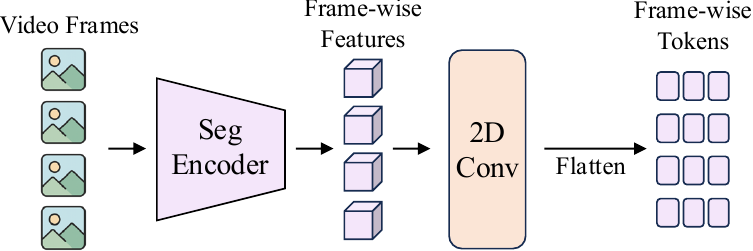} 
    \caption{Frame-Aware Video Tokenizer architecture.} 
    \label{fig:video-tokenizer}
\end{figure}

As shown in Fig.~\ref{fig:video-tokenizer}, we extract frame-wise segmentation features using the vision encoder of the segmentation model:
\begin{equation}
\mathbf{S}_{\text{seg}} \in \mathbb{R}^{H' \times W' \times T_{\text{seg}} \times C}.
\end{equation}

Each feature map is downsampled through three $3{\times}3$ stride-2 convolutions. 
In our setting, $H' = 64$, $W'=64$ and $C = 256$, so applying three successive $\tfrac{1}{2}$-scale convolutions reduces the spatial size to $H'/8 = 8$:
\begin{equation}
\mathbf{S}_{\text{down}} \in
\mathbb{R}^{8 \times 8 \times T_{\text{seg}} \times C}.
\end{equation}

The $8{\times}8$ grid is flattened into $64$ spatial tokens per frame and projected into $D$ dimensions:
\begin{equation}
\mathbf{S}_{\text{patch}}
= \mathrm{Linear}\!\left(\mathrm{reshape}(\mathbf{S}_{\text{down}})\right)
\in \mathbb{R}^{T_{\text{seg}} \times 64 \times D}.
\end{equation}

Finally, the temporal and spatial axes are merged to construct the video-token sequence used in the cross-attention mechanism with $\mathbf{E}_{\text{ST}}\in \mathbb{R}^{N\times D}$:
\begin{equation}
\mathbf{S}_{\text{vid}}
\in \mathbb{R}^{(T_{\text{seg}} \times 64) \times D}.
\end{equation}

\section{Efficiency Analysis}
\label{supp:efficiency}

\subsection{Effect of $\alpha$ on Performance and Efficiency}
We analyze the effect of $\alpha$ on performance and efficiency. As shown in Tab.~\ref{tab:ablation_alpha_efficiency} and Tab.~\ref{tab:ablation_num_anchor_frames}, increasing $\alpha$ improves performance with only minor impact on efficiency. Peak memory remains nearly constant (within $<0.01$ GB), while FPS decreases slightly (from 5.14 to 4.98 as $\alpha$ increases from 2 to 6).

\begin{table}[t]
\centering
\footnotesize
\setlength{\tabcolsep}{6pt}
\renewcommand{\arraystretch}{1.0}
\caption{Efficiency of \virstabbr{} with different $\alpha$.}
\label{tab:ablation_alpha_efficiency}
\begin{tabular}{ccc}
\toprule
\multirow{2}{*}{\textbf{$\alpha$}} & \multicolumn{2}{c}{\textbf{Efficiency}} \\
\cmidrule(lr){2-3}
 & FPS$\uparrow$ & Memory (GB)$\downarrow$ \\
\midrule
2 & 5.14 & 37.52 \\
4 & 5.04 & 37.52 \\
6 & 4.98 & 37.52 \\
\bottomrule
\end{tabular}
\end{table}

\subsection{Inference Efficiency Comparison}
We compare the inference efficiency of \virstabbr{} with existing methods in Tab.~\ref{tab:supp_time_efficiency}. We report FPS on the MeViS dataset using a single A100 GPU. \virstabbr{} achieves 5.10 FPS, compared to 3.81 FPS for VRS-HQ-7B and 1.47 FPS for VISA-7B. These results indicate that \virstabbr{} maintains competitive efficiency while performing joint reasoning and segmentation.

\begin{table}[t]
\centering
\footnotesize
\setlength{\tabcolsep}{6pt}
\renewcommand{\arraystretch}{1.0}
\caption{Inference speed comparison across methods.}
\label{tab:supp_time_efficiency}
\begin{tabular}{lc}
\toprule
Method & FPS$\uparrow$ \\
\midrule
VISA-7B~\cite{yan2024visa_forrebuttal} w/o postproc. & 1.47 \\
VRS-HQ-7B~\cite{gong2025vrshq_forrebuttal} & 3.81 \\
HyperSeg-3B~\cite{Wei_2025_hyperseg_forrebuttal} & 1.54 \\
VIRST-7B ($\alpha=3$, Ours) & \textbf{5.10} \\
\bottomrule
\end{tabular}
\end{table}

\section{Qualitative Results}
\subsection{Video Segmentation Qualitative Results}
\label{supp:video_qualitative}
\begin{figure*}[t]
  \centering

    \includegraphics[width=1\linewidth]{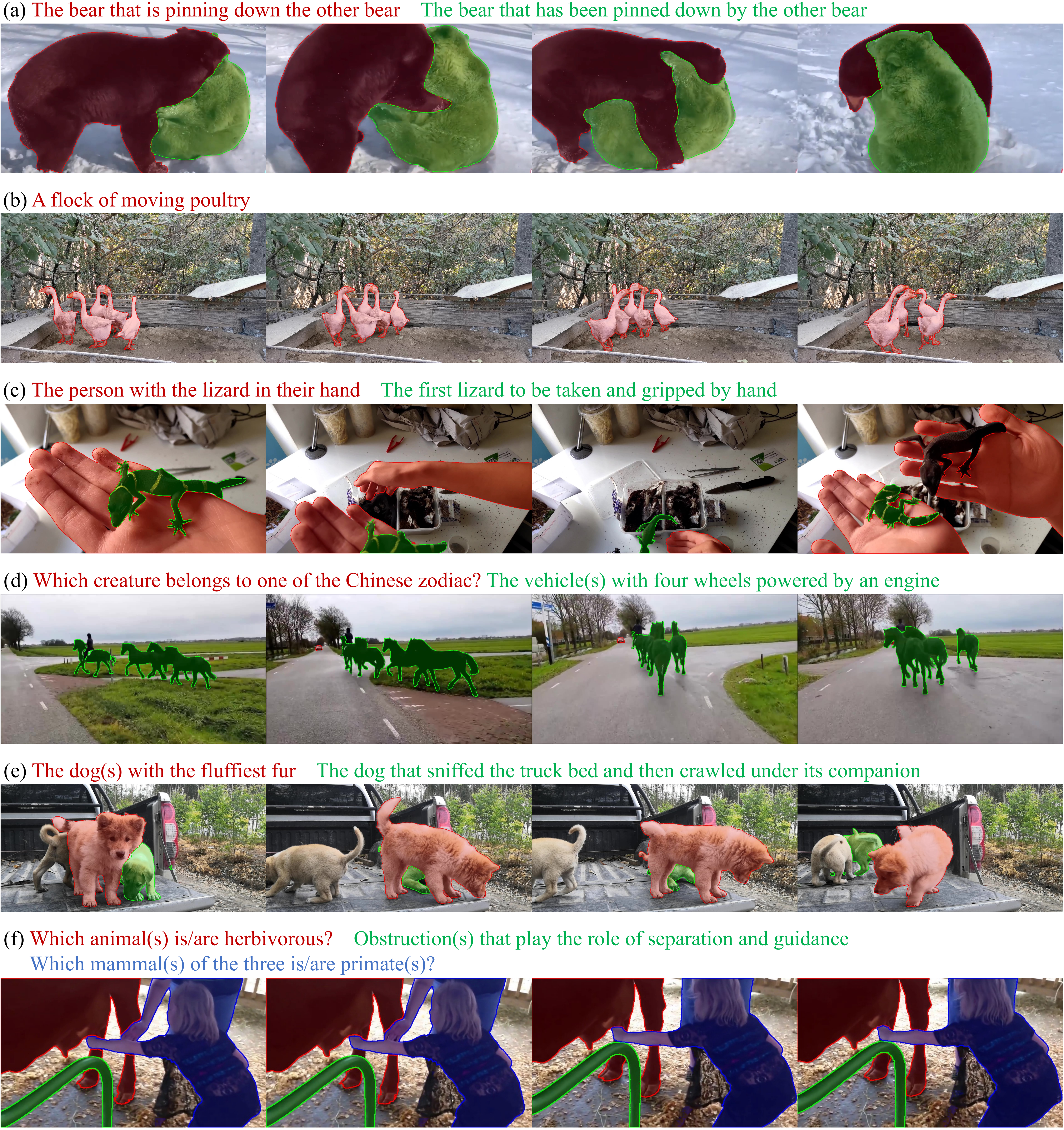} 
    \caption{
\textbf{Qualitative results of \virstabbr\ on videos.}
Results are best viewed when zoomed in.
    }
    \label{fig:supp_vid_qualitative_results}
\end{figure*}

Fig.~\ref{fig:supp_vid_qualitative_results} provides additional qualitative results for video segmentation, demonstrating strong performance across challenging cases such as multi-object scenes, heavy distractors, and small-object targets.

\subsection{Image Segmentation Qualitative Results}
\label{supp:image_qualitative}
\begin{figure*}[t]
  \centering

    \includegraphics[width=1\linewidth]{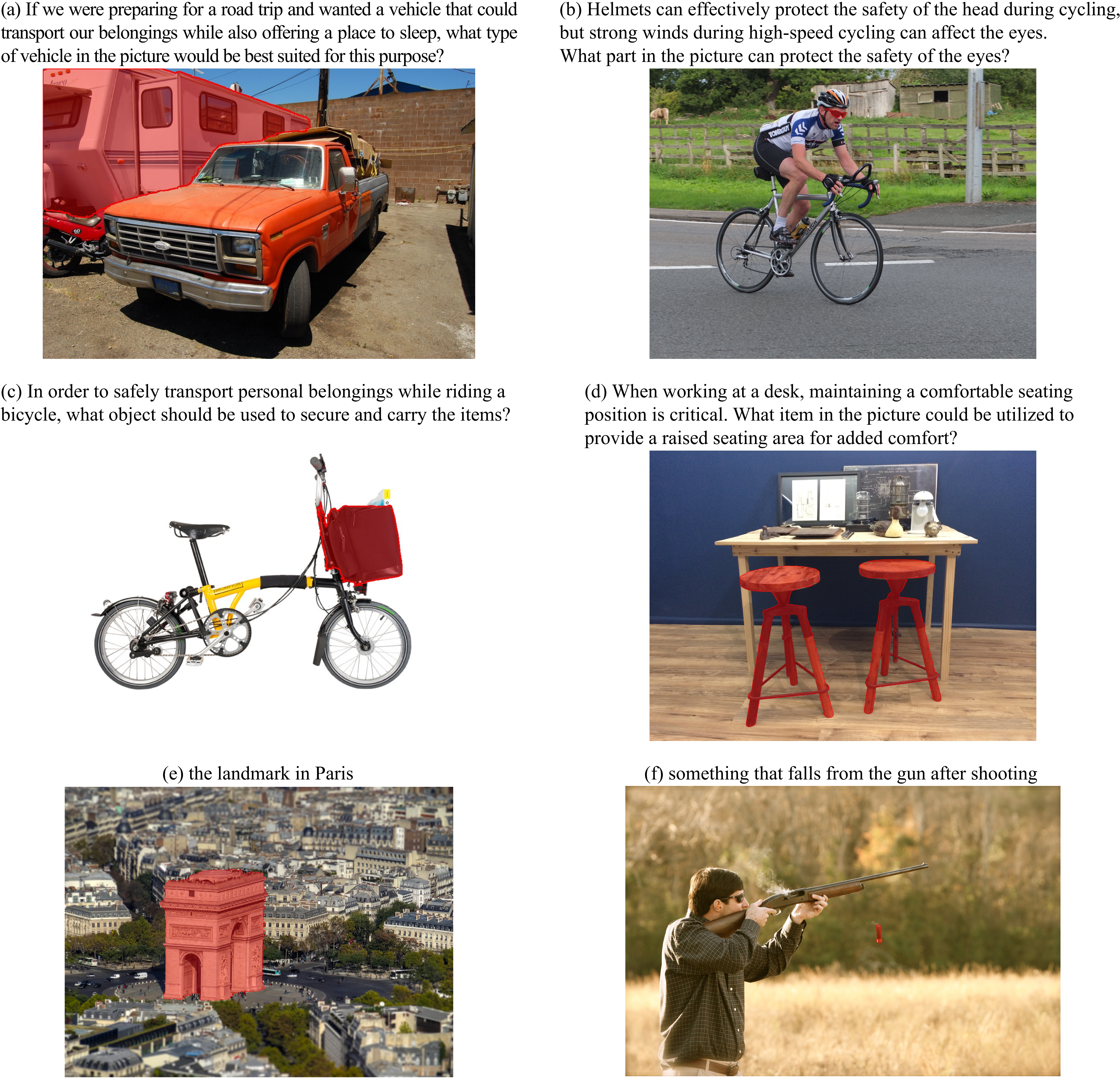} 
    \caption{
\textbf{Qualitative results of \virstabbr\ on images.}
Results are best viewed when zoomed in.
    }
    \label{fig:supp_image_qualitative_results}
\end{figure*}

Fig.~\ref{fig:supp_image_qualitative_results} presents additional image reasoning-segmentation results, showing that the model can accurately localize objects even under complex, fine-grained textual descriptions.

\subsection{Video Understanding Qualitative Results}
\label{supp:video_understanding}
\begin{figure*}[t]
  \centering

    \includegraphics[width=1\linewidth]{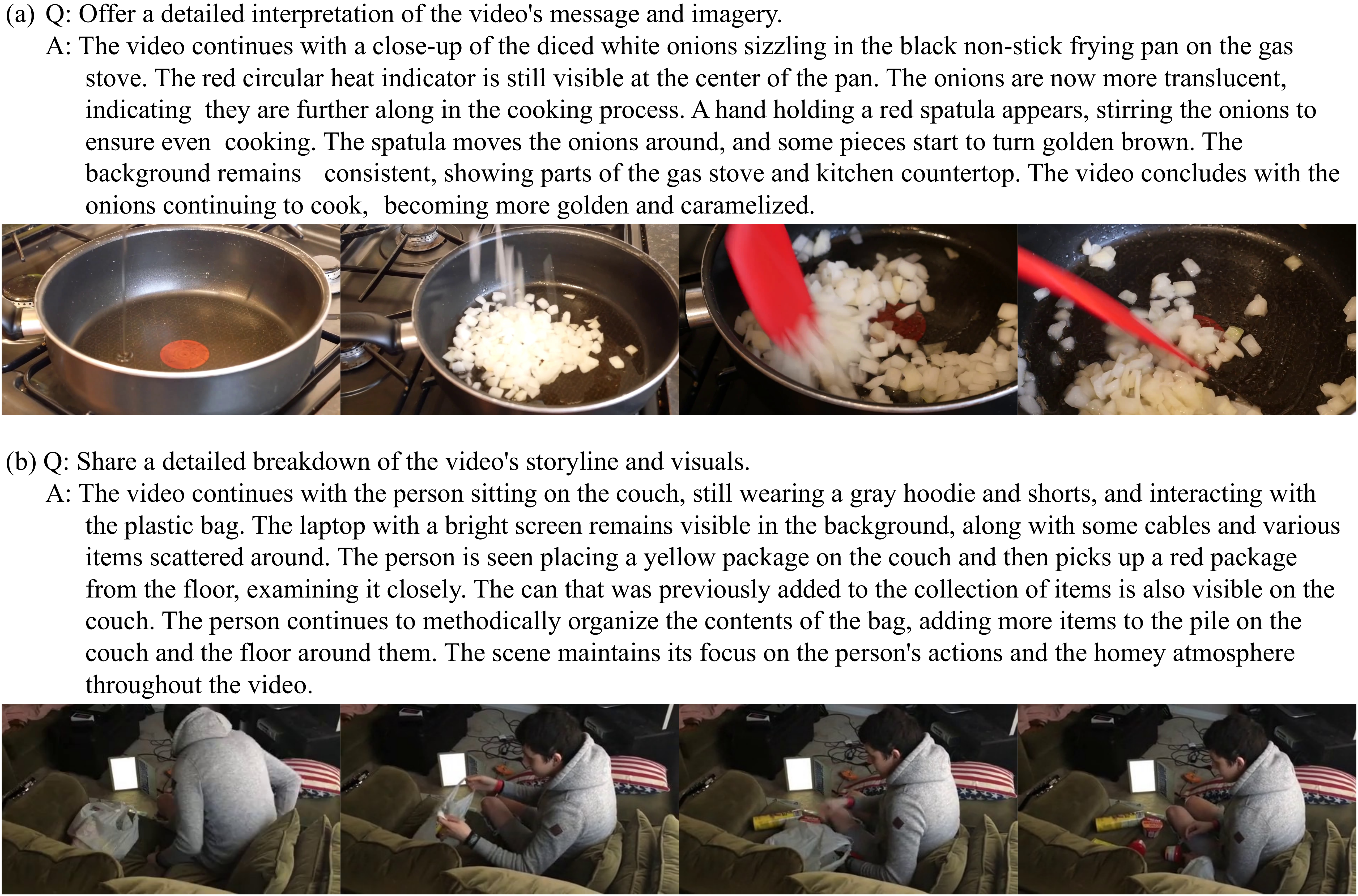} 
    \caption{
    \textbf{Qualitative video understanding results of \virstabbr{}.}
    }
    \label{fig:supp_vqa}
\end{figure*}

Fig.~\ref{fig:supp_vqa} demonstrates that \virstabbr{} retains strong video understanding capability, with responses generated autoregressively.

\subsection{Failure Cases}

\label{supp:failure_cases}

\begin{figure*}[t]
  \centering
    \includegraphics[width=1\linewidth]{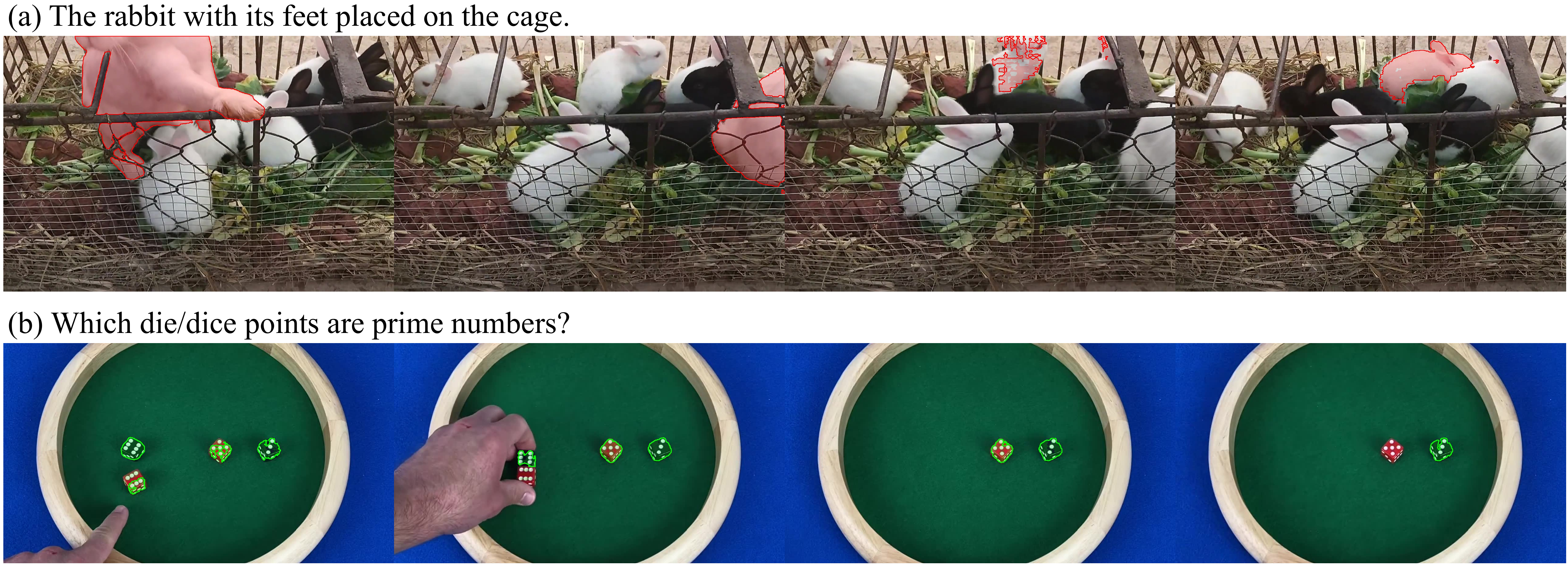} 
    \caption{
    \textbf{Failure cases of \virstabbr{}.}    }
    \label{fig:supp_failure}
\end{figure*}

Fig.~\ref{fig:supp_failure} shows failure cases of \virstabbr{}.
In Fig.~\ref{fig:supp_failure} (a), the scene contains many visually similar distractors, making the scenario inherently difficult. Although the queried object and its defining motion appear in the first frame, the target moves rapidly and undergoes heavy occlusions. \virstabbr{} initially tracks it, but the mask gradually drifts toward a similar distractor and eventually switches to it.

Fig.~\ref{fig:supp_failure} (b) requires multi-step semantic reasoning: the task is to segment only the dice showing 3 and 5 (prime numbers). \virstabbr{} struggles to maintain this constraint over time, intermittently masking the die showing 6 and failing to consistently retain the mask for 5 in the final frame.

\section{Limitations and Future Directions}

While \virstabbr{} demonstrates strong performance in complex scenes and reasoning-intensive queries, several limitations remain. As shown in Appendix~\ref{supp:failure_cases}, the model still struggles in highly cluttered environments with many distractor objects, and it can fail when the query requires multi-step semantic reasoning. Future work should explore training strategies that more explicitly ground step-by-step reasoning in video inputs, enabling tighter integration between pixel-level visual understanding and compositional language reasoning, and allowing VLMs to extend more effectively to long-video segmentation and complex video scenarios.

\end{document}